\RequirePackage{rotating}

\documentclass{new_tlp}
\usepackage{mathptmx}

\usepackage[most]{tcolorbox}
\usepackage{makeidx}
\usepackage[english]{babel}
\usepackage{amsmath}
\usepackage{amsfonts}
\usepackage{amssymb}
\usepackage{graphicx}
\usepackage{xspace}
\usepackage{url}
\usepackage{algorithm}
\usepackage{algpseudocode}
\usepackage{booktabs}
\usepackage{multirow}
\usepackage[labelsep=quad,indention=10pt]{subfig}
\usepackage{bbding}
\usepackage{todonotes}
\usepackage{adjustbox}
\usepackage{pgfplots}
\usepackage{rotating}
\usepackage{paralist}

\newcommand{\sysfont}{\textit}
\newcommand{\lpopt}{\sysfont{lpopt}\xspace}

\newcommand{\wasp}{\sysfont{wasp}\xspace}
\newcommand{\clasp}{\sysfont{clasp}\xspace}

\newcommand{\dlv}{\sysfont{DLV}\xspace}
\newcommand{\idlv}{{{\small $\cal I$}-}\dlv}
\newcommand{\idlvsd}{{{{\small $\cal I$}-}\sysfont{DLV}}$^{\textsc{sd}}$\xspace}

\newcommand{\aspcore}{$\mathsf{ASP\text{-}Core\text{-}2}$\xspace}
\newcommand{\derives}{\mbox{\,:\hspace{0.1em}\texttt{-}}\,\xspace}

\newcommand{\Or}{\ensuremath{\ |\ }\xspace}
\newcommand{\p}{\ensuremath{{P}}\xspace}
\newcommand{\GP}{\emph{Ground(\p)}}
\newcommand{\naf}{\ensuremath{not\ }\xspace}

\newcommand{\decalg}{\textsc{{SmartDecomposition}}\xspace}

\newcommand{\anonym}{\text{\underline{\hspace{0.2cm}}}}
\def\Facts{\emph{Facts}}
\def\EDB{\emph{EDB}}
\def\IDB{\emph{IDB}}
\newcommand{\BP}{\ensuremath{B_{\p}}\xspace}
\newcommand{\UP}{\ensuremath{U_{\p}}\xspace}

\newcommand{\cit}[1]{~\cite{#1}}
\newcommand{\re}[1]{~\ref{#1}}

\newcommand{\commentsymbolright}{/*}
\newcommand{\commentsymbolleft}{*/}
\algrenewcommand\algorithmiccomment[1]{\hfill \commentsymbolright{} #1 \commentsymbolleft{}}
\makeatletter
\newcommand{\LineComment}[2][\algorithmicindent]{\State \hspace{#1}\commentsymbolright{} #2 \commentsymbolleft{}}
\makeatother

\newenvironment{dlvcode}
  {\begin{displaymath}\begin{array}{l}}
  {\end{array}\end{displaymath}}

\newtheorem{example}{Example}

\begin{document}
\bibliographystyle{acmtrans}

\title[Optimizing Answer Set Computation via Heuristic-Based Decomposition]
    {Optimizing Answer Set Computation via Heuristic-Based Decomposition~\footnote{This work is the extended version of a paper originally appeared in the Proceedings of 20th Symposium on Practical Aspects of Declarative Languages (PADL 2018), January 8--9, 2018, Los Angeles, USA. Program chairs were Kevin Hamlen and Nicola Leone.
    The paper presents new material that integrates and extends what has been reported in the original paper; in particular, it provides the reader with proper preliminaries (omitted in the original paper for space constraints), more detailed discussions on the proposed techniques and richer comparisons with related approaches, along with an extended number of examples.
    Furthermore, a more thorough experimental activity is presented, discussed in part in the main text and in part in the appendices, that covers also new domains that were unexplored in the original paper.}
    }

\author[F. Calimeri, S. Perri and J. Zangari]
    {Francesco Calimeri, Simona Perri and Jessica Zangari\\
    	Department of Mathematics and Computer Science, University of Calabria, \\
        Rende, Italy \\
    	\email{$\{$calimeri,perri,zangari$\}$@mat.unical.it}
    }

\maketitle

\begin{abstract}

Answer Set Programming (ASP) is a purely declarative formalism developed in the field of logic programming and nonmonotonic reasoning: computational problems are encoded by logic programs whose answer sets, corresponding to solutions, are computed by an ASP system.
Different, semantically equivalent, programs can be defined for the same problem; however, performance of systems evaluating them might significantly vary.
We propose an approach for automatically transforming an input logic program into an equivalent one that can be evaluated more efficiently.
One can make use of existing tree-decomposition techniques for rewriting  selected rules into a set of multiple ones; the idea is to guide and adaptively apply them on the basis of proper new heuristics, to obtain a smart rewriting algorithm to be integrated into an ASP system.
The method is rather general: it can be adapted to any system and implement different preference policies.
Furthermore, we define a set of new heuristics tailored at optimizing grounding, one of the main phases of the ASP computation; we use them in order to implement the approach into the ASP system \dlv, in particular into its grounding subsystem \idlv, and carry out an extensive experimental activity for assessing the impact of the proposal.
\end{abstract}

%
%
\section{Introduction}\label{sec:intro}
Answer Set Programming (ASP)~\cite{DBLP:journals/cacm/BrewkaET11,DBLP:journals/ngc/GelfondL91} is a declarative programming paradigm proposed in the area of non-monotonic reasoning and logic programming.
With ASP, computational problems are encoded by logic programs whose answer sets, corresponding to solutions, are computed by an ASP system~\cite{DBLP:conf/iclp/Lifschitz99}.

The evaluation of ASP programs is ``traditionally'' split into two phases: \emph{grounding}, that generates a propositional theory semantically equivalent to the input program, and \emph{solving}, that applies  propositional techniques for computing the intended semantics~\cite{DBLP:conf/lpnmr/AlvianoCDFLPRVZ17,DBLP:conf/lpnmr/GebserKK0S15,DBLP:journals/aim/KaufmannLPS16,DBLP:journals/tocl/LeonePFEGPS06};
nevertheless, in the latest years several approaches that deviate from this schema have been proposed~\cite{DBLP:journals/fuin/PaluDPR09,DBLP:conf/jelia/Dao-TranEFWW12,DBLP:conf/ijcai/EiterKW17,DBLP:journals/tplp/LefevreBSG17}.

Typically, the same computational problem can be encoded by means of many different ASP programs which are semantically equivalent; however, real ASP systems may perform very differently when evaluating each one of them.
This behavior is due, in part, to specific aspects, that strictly depend on the ASP system  employed, and, in part, to general ``intrinsic'' aspects, depending on the program at hand which could feature some characteristics that can make computation easier or harder.
Thus, often, to have satisfying performance, expert knowledge is required in order to
select the best encoding.
This issue, in a certain sense, conflicts with the declarative nature of ASP that, ideally, should free the users from the burden of the computational aspects. For this reason, ASP systems tend to be endowed with proper pre-processing means aiming at making performance less encoding-dependent; intuitively, this is crucial for fostering and easing the usage of ASP in practice.

A proposal in this direction is \lpopt\cit{DBLP:conf/lopstr/BichlerMW16}, a pre-processing tool for ASP systems that rewrites rules in input programs by means of \textit{tree-decomposition} algorithms.
The rationale comes from the fact that, when programs contain rules featuring long bodies, ASP systems performance might benefit from a careful split of such rules into multiple, smaller ones.
However, it is worth noting that, while in some cases such decomposition is
convenient, in other cases keeping the original rule is preferable; hence, a black-box
decomposition, like the one of \lpopt, makes it difficult to predict whether it will lead to benefits or disadvantages.

Inspired by the idea implemented in \lpopt of rewriting ASP programs by means of tree-decomposition, we propose here a method that aims at taking full advantage from rewriting, still avoiding performance drawbacks by estimating its effects in advance.
It analyzes each input rule before the evaluation, and estimates whether it is convenient to decompose it into an equivalent set of smaller rules, or not; if more than one decomposition is possible, the most promising is selected.
The method is general and defined so that all choices are made according to proper criteria and heuristics that can be customized: it can be tailored to different phases of the ASP computation, and it is not tied to a specific system.
Furthermore, we define new heuristics and criteria relying on data and statistics dynamically computed during the instantiation with the aim of optimizing the performances of \idlv~\cite{DBLP:journals/ia/CalimeriFPZ17}, a recently released deductive database system that currently serves also as the
grounding subsystem of \dlv\cit{DBLP:conf/lpnmr/AlvianoCDFLPRVZ17}.
In addition, we present here an actual implementation into \idlv and perform
an extensive experimental activity in order to asses the effects of our technique on ASP program optimization.

\medskip

The remainder of the paper is structured as follows.
In Section~\ref{sec:preliminaries} we recall ASP basics along with some other preliminary notions; in Section~\ref{sec:general} we introduce an abstract heuristic-guided decomposition algorithm for ASP programs in its general form, while in Section~\ref{subsec:specdecalg} we describe how we adapt it in order to foster an actual implementation into the \idlv grounder, along with custom heuristics for guiding the process.
Section~\ref{sec:experiments} presents the results of an extensive experimental activity aimed at assessing the impact of the proposed method, and effectiveness of the proposed heuristics on grounding performance; we also shed a light on the impact on solvers.
Our conclusions are drawn in Section~\ref{sec:conclusion}.
Some additional experiments, that have been omitted from the main text for the sake of readability, are reported and discussed in appendices.

%
%
\section{Preliminaries}\label{sec:preliminaries}

In this section we provide the reader with some preliminaries; in particular, we first briefly introduce Answer Set Programming and then recall how hypergraphs can be used in order to represent ASP rules along with tree-decomposition strategies for rewriting them.

\subsection{Answer Set Programming}\label{subsec:ASP}

A significant amount of work has been carried out on extending the basic language of ASP, and the community recently agreed on a standard input language for ASP systems: \aspcore~\cite{asp-core-2-01c}, the official language of the ASP Competition series~\cite{DBLP:journals/ai/CalimeriGMR16,DBLP:conf/aaai/GebserMR16}.
For the sake of simplicity, we focus next on the basic aspects of the language; for a complete reference to the \aspcore standard, and further details about advanced ASP features, we refer the reader to~\cite{asp-core-2-01c} and the vast literature.

\ A \emph{term} is either a \emph{simple term} or a \emph{functional term}.
A {\em simple term} is either a constant or a variable. If $t_1 \ldots t_n$ are terms and $f$ is a function symbol of arity $n$, then $f(t_1, \dots, t_n)$ is a {\em functional term}. If $t_1,\dots, t_k$ are terms and $p$ is a \emph{predicate symbol} of arity $k$, then $p(t_1,\dots, t_k)$ is an {\em atom}.
A {\em literal} $\,l$ is of the form $a$ or $\naf a$, where $a$ is an atom; in the former case $l$ is {\em positive}, otherwise {\em negative}.
A {\em rule $r$} is of the form\ \ $ \alpha_1\Or\cdots\Or\alpha_k \derives $ $ \beta_1, \dots, \beta_n,$ $\naf\, \beta_{n+1},\dots,\naf\, \beta_{m}. $\ \ where $m\geq 0$, $k \geq 0$; $\alpha_1,\ldots,\alpha_k$ and $\beta_1, \dots, \beta_{m}$ are atoms. We define $H(r)$ = $\{ \alpha_1,$ $\ldots,$ $\alpha_k\}$ (the {\em head} of $r$) and $B(r) = B^+(r) \cup B^-(r)$ (the {\em body} of $r$), where $B^+(r) = \{ \beta_1,$ $\dots,$ $\beta_n\}$ (the {\em positive body}) and $B^-(r)$ = $\{\naf\ \beta_{n+1},$ $\dots,$ $\naf\ \beta_{m} \}$ (the {\em negative body}).
If $H(r) = \emptyset$ then $r$ is a {\em (strong) constraint}; if $B(r) =\emptyset$ and $|H(r)| =1$ then $r$ is a {\em fact}.
A rule $r$ is safe if each variable of $r$ has an occurrence in $B^+(r)$\footnote{We remark that this definition of safety is specific for the syntax considered herein.
For a complete definition we refer the reader to~\cite{asp-core-2-01c}.}.
For a rule $r$, we denote as $headvar(r)$, $bodyvar(r)$ and $var(r)$ the set of variables occurring in $H(r)$, $B(r)$ and $r$, respectively.
An ASP program is a finite set $P$ of safe rules.
A program (a rule, a literal) is {\em ground} if it contains no variables.
A predicate is defined by a rule $r$ if it occurs in $H(r)$. A predicate defined only by facts is an \EDB\ predicate, the remaining are \IDB\ predicates. The set of all facts in $P$ is denoted by \Facts($P$); the set of instances of all \EDB\ predicates in $P$ is denoted by \EDB($P$).

Given a program $P$, the {\em Herbrand universe} of $P$, denoted by \UP, consists of all ground terms that can be built combining constants and function symbols appearing in $P$.
The {\em Herbrand base} of~$P$, denoted by \BP, is the set of all ground atoms obtainable from the atoms of $P$ by replacing variables with elements from \UP. A {\em substitution} for a rule $r \in P$ is a mapping from the set of variables of $r$ to the set \UP of ground terms.
A {\em ground instance} of a rule $r$ is obtained applying a substitution to $r$. The {\em full instantiation} \GP\ of $P$ is defined as the set of all ground instances of its rules over \UP. An {\em interpretation} $I$ for $P$ is a subset of \BP. A positive literal $a$ (resp., a negative literal $\naf\ a$) is true w.r.t. $I$ if $a \in I$ (resp., $a \notin I$); it is false otherwise. Given a ground rule $r$, we say that $r$ is satisfied w.r.t. $I$ if some atom appearing in $H(r)$ is true w.r.t. $I$ or some literal
appearing in $B(r)$ is false w.r.t. $I$. Given a program $P$, we say that $I$ is a {\em model} \ of $P$, iff all rules in \GP\ are satisfied w.r.t. $I$. A model $M$ is {\em minimal} if there is no model $N$ for $P$ such that $N \subset M$. The {\em Gelfond-Lifschitz reduct}~\cite{DBLP:journals/ngc/GelfondL91} of $P$, w.r.t.\ an interpretation $I$, is the positive ground program $P^I$ obtained from $\GP$ by: $(i)$ deleting all rules having a negative literal false w.r.t. $I$; $(ii)$ deleting all negative literals from the remaining rules. $I\subseteq \BP$ is an {\em answer set} for a program $P$ iff
$I$ is a minimal model for $P^I$.
The set of all answer sets for $P$ is denoted by $AS(P)$.


\subsection{ASP Computation}\label{subsec:asp-computation}
The high expressiveness of ASP comes at the price of a high computational cost in the worst case~\cite{DBLP:journals/tods/EiterGM97,DBLP:journals/tocl/LeonePFEGPS06}, which makes the implementation of efficient ASP systems a difficult task.
Thanks to the effort by a scientific community that grew over time, there are nowadays a number of systems that support ASP and its variants~\cite{DBLP:journals/ai/SimonsNS02,DBLP:conf/lpnmr/WardS04,DBLP:journals/tocl/JanhunenNSSY06,DBLP:journals/jar/GiunchigliaLM06,DBLP:journals/ai/GebserKS12,DBLP:conf/lpnmr/AlvianoDLR15,DBLP:journals/tocl/LeonePFEGPS06,DBLP:journals/corr/GebserKKS14,DBLP:journals/fuin/PaluDPR09,DBLP:journals/tplp/LefevreBSG17,DBLP:conf/jelia/Dao-TranEFWW12,DBLP:conf/lpnmr/Weinzierl17}.

The well-established, mainstream approach for the evaluation of ASP programs\cit{DBLP:journals/aim/KaufmannLPS16} relies on two phases, usually referred to as {\em instantiation} or {\em grounding}, and {\em solving} or {\em answer sets search}, respectively.
Given a (non-ground) ASP program $P$, grounding consists of producing a propositional theory $G_{P}$ semantically equivalent to $P$, i.e. such that $G_P$ does not contain any variable but has the same answer sets as $P$. 
Given that, in the worst case, the solving stage may take up to exponential time in the size of $G_P$~\cite{DBLP:journals/amai/Ben-EliyahuD94,DBLP:journals/ai/Ben-Eliyahu-ZoharyP97}, modern ASP systems employ intelligent grounding procedures so that $G_P$ is significantly smaller than the full instantiation \GP. Once the program $G_P$ has been computed, solving takes place, taking as input $G_{P}$ and computing its answer sets by means of propositional algorithms.
The majority of current ASP implementations follows this two-phase computation, either by explicitly relying on stand-alone grounders~\cite{DBLP:conf/lpnmr/Syrjanen01,DBLP:conf/birthday/FaberLP12,DBLP:conf/lpnmr/GebserKKS11} and solvers~\cite{DBLP:journals/ai/SimonsNS02,DBLP:conf/lpnmr/WardS04,DBLP:journals/tocl/JanhunenNSSY06,DBLP:journals/jar/GiunchigliaLM06,DBLP:journals/ai/GebserKS12,DBLP:conf/lpnmr/AlvianoDLR15}, or integrating the modules into monolithic systems~\cite{DBLP:journals/corr/GebserKKS14,DBLP:journals/tocl/LeonePFEGPS06,DBLP:conf/lpnmr/AlvianoCDFLPRVZ17}.
%
Notably, given that both phases are, in general, computationally expensive~\cite{DBLP:journals/tods/EiterGM97,DBLP:journals/csur/DantsinEGV01}, efficient ASP implementations depend on proper optimization of both.

Alternative solutions~\cite{DBLP:journals/fuin/PaluDPR09,DBLP:journals/tplp/LefevreBSG17,DBLP:conf/jelia/Dao-TranEFWW12,DBLP:conf/lpnmr/Weinzierl17} adopt a \textit{lazy grounding} technique, in which grounding and solving steps
are interleaved, and rules are grounded on-demand during solving.
These systems try to overcome the so called \textit{grounding bottleneck}, that occurs on problems for which the instantiation is inherently so huge that its actual materialization is not suitable in practice.
For this reason, this approach looks promising; however, current implementations do not match, in the general case, performance of the more ``traditional'' systems, that proved instead to be reliable and well-performing in a wider range of scenarios.

Notably, the herein presented technique, introduced in Section~\ref{sec:general}, is general enough to be adopted with both approaches, by defining suitable heuristics and properly customizing its integration.


\subsection{Tree-Decompositions for Rewriting ASP Rules}\label{subsec:TreeDecomp}
Hypergraphs are useful for describing the structure of many computational problems; furthermore, it is possible to decompose them into different parts, so that the solution(s) of problems can be obtained by a polynomial divide-and-conquer algorithm that properly exploits this division~\cite{DBLP:conf/mfcs/GottlobLS01,DBLP:conf/wg/GottlobGMSS05}.
Such ideas can guide a rewriting of an ASP program: indeed, a logic rule can be represented as a \textit{hypergraph}~\cite{DBLP:conf/iclp/MorakW12}, and hence properly decomposed.

Discussing in detail how tree decompositions can be computed and rewritings induced is out of the scope of this paper; indeed, our main goal is to find a way for correctly identifying in advance in which cases their application pays off in terms of efficiency, while dealing with ASP rules.
However, in order to ease the reading, in the following, we briefly recall an intuitive description of some crucial notions for the ASP context; for further details we refer the reader to\cit{DBLP:conf/lopstr/BichlerMW16} and the existing literature.

\medskip

A (undirected) hypergraph is a generalization of a (undirected) graph in which an edge can join two or more vertices.
An ASP rule $r$ can be represented as a hypergraph $HG(r)$ by adding a hyperedge for each literal $l \in B(r)\cup H(r)$ containing the variables appearing in $l$.
A tree decomposition of a hypergraph $HG(r)$ (see~\cite{DBLP:journals/jal/RobertsonS86,DBLP:conf/pods/GottlobGLS16}) is a tuple ($TD(r)$, $\chi$), where $TD(r)$ = $(V(TD(r)), E(TD(r)))$ is a tree and $\chi :$ $V(TD(r)) \rightarrow$ $2^{V(HG(r))}$ is a function associating a set of vertices $\chi(t) \subseteq V(HG(r))$ to each vertex $t$ of the decomposition tree $TD(r)$, such that for each $e \in E(HG(r))$ there is a node $t \in V(TD(r))$ such that $e \subseteq \chi(t)$, and for each $v \in  V(HG(r))$ the set $\{t \in V(TD(r)) | v \in \chi(t)\}$ is connected in $TD(r)$.
Intuitively, a tree decomposition $TD(r)$ of $HG(r)$ is a tree such that each vertex is associated to a \emph{bag}, i.e., a set of nodes of $HG(r)$, and such that each hyperedge of $HG(r)$ is covered by some bag, and for each node of $HG(r)$ all vertices of $TD(r)$ whose bag contains it induce a connected subtree of $TD(r)$.
%

A tree decomposition $TD(r)$ can be used in order to produce a set of rules that rewrites $r$; such set is called {\em rule decomposition}, and denoted by $RD(r)$.
In particular, $RD(r)$ contains a (newly generated) rule for each vertex $v$ of $TD(r)$, on the basis of the included variables.
Roughly, each literal $l$ in the body of $r$, such that the set of variables in $l$ is contained in $v$, is added to the body of the rule generated for $v$.
Eventually, some optional rules may be added to $RD(r)$ in order to guarantee safety. Note that, since different choices for handling safety can be performed, the way in which a tree decomposition is converted into a rule decomposition might be not unique. Moreover, interestingly, in general, more than one decomposition is possible for each rule.

The following running example, which we will refer to throughout the paper, illustrates this mechanism.

\begin{example}\label{ex:nomystery}
Let us consider the rule: 
$$r_1: p(X,Y,Z,S)\derives s(S),\ a(X,Y,S-1),\ c(D,Y,Z),\
f(X,P,S-1),\ P >= D.$$
from the encoding of the problem \textit{Nomystery} from the $6th$ ASP
Competition (see Section~\ref{sec:experiments}), where, for the sake of readability, predicates and variables have been renamed.
Figure\re{fig:nmi} depicts the conversion of $r_1$ into the hypergraph $HG(r_1)$, along with two
possible decompositions: $TD_1(r_1)$ and $TD_2(r_1)$, that induce two different rewritings.
According to $TD_1(r_1)$, $r_1$ can be rewritten into the set of rules $RD_1(r_1)$:

\begin{figure}[t]
\begin{small}
\begin{tcolorbox}
    \centering
       \subfloat[$HG(r_1)$]{%
          \includegraphics[width=28.0mm]{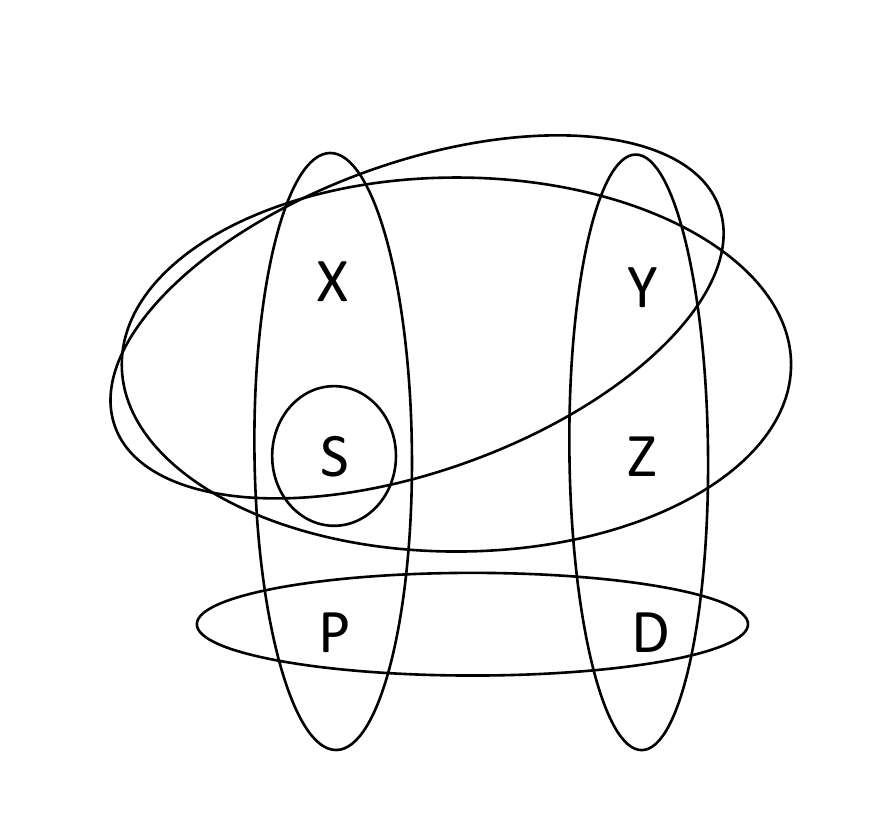}%
       }
       \subfloat[$TD_1(r_1)$]{%
          \includegraphics[width=25.0mm]{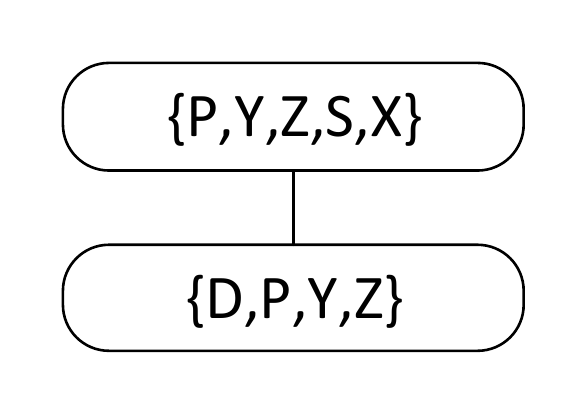}%
       }
       \subfloat[$TD_2(r_1)$\vspace{2.1ex}]{%
          \includegraphics[width=25.0mm]{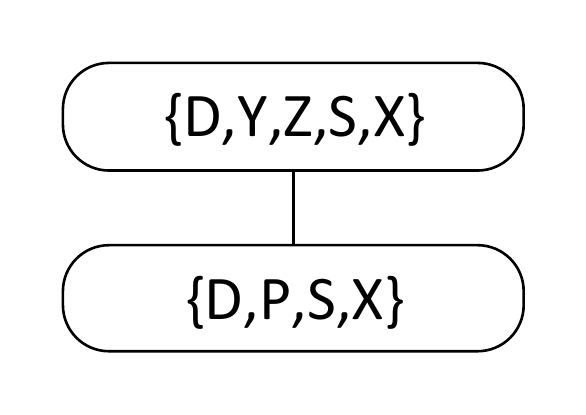}%
       }
       \caption{\small Decomposing a rule \label{fig:nmi} } 
\end{tcolorbox}
\end{small}
\end{figure}

\begin{dlvcode}
    r_2: p(X,Y,Z,S) \derives s(S), a(X,Y,S-1), f(X,P,S-1), fresh\_pred\_1(P,Y,Z).\\
    r_3: fresh\_pred\_1(P,Y,Z) \derives c(D,Y,Z), P>=D, fresh\_pred\_2(P).\\
    r_4: fresh\_pred\_2(P) \derives s(S), f(\anonym,P,S-1).
\end{dlvcode}


In particular, the rule $r_2$ features the same head of $r_1$ and as body the literals needed in order to cover the node of $TD_1(r_1)$ containing the variables $\{P,Y,Z,S,X\}$; $r_3$ features as head the fresh predicate $fresh\_pred\_1$ that links it to $r_2$ and collects in its body a set of literals covering the variables $\{D,P,Y,Z\}$ appearing in the other node of $TD_1(r_1)$; eventually, $r_4$ is needed to ensure safety of $r_3$: the atom $fresh\_pred\_2(P)$ is added in the body of $r_3$ and to the head of $r_4$, whose body features a set of literals coming from $r_1$ and covers $P$ (note that in this case the set is unique). Note that, a different rewriting could be obtained by differently handling safety of $r_3$; for instance, one could avoid to introduce $r_4$ and, instead, add the literals  $s(S)$, and $f(\anonym,P,S-1)$ to the body of $r_3$.

Similarly, according to $TD_2(r_1)$, $r_1$ can be rewritten into $RD_2(r_1)$ as follows:

\begin{dlvcode}
    r_5: p(X,Y,Z,S) \derives a(X,Y,S-1),c(D,Y,Z), fresh\_pred\_1(D,S,X).\\
    r_6: fresh\_pred\_1(D,S,X) \derives s(S),f(X,P,S-1),P>=D,fresh\_pred\_2(D).\\
    r_7: fresh\_pred\_2(D) \derives c(D,\anonym,\anonym).
\end{dlvcode}
\end{example}


\section{A Heuristic-guided Decomposition Algorithm}\label{sec:general}
In the previous section we recalled how tree decomposition of hypergraphs can be used in order to guide rewritings of ASP rules.
Interestingly, the \lpopt\cit{DBLP:conf/lopstr/BichlerMW16} preprocessor is a proposal in this direction, that rewrites an ASP program before it is fed to an ASP system.

As previously noted, for each rule, several different rule decompositions might exist.
However, when fed to a real ASP system, different yet equivalent rewritings require, in general, significantly different evaluation times.
Thus, proper means for reasonably and effectively choose the ``best'' rewriting are crucial; furthermore, it might be the case that, whatever the choice, sticking to the original, unrewritten rule, is still preferable.
Hence, a black-box approach, such as the one of \lpopt, makes it difficult to
effectively take advantage from the decomposition rewritings; this is clearly noticeable by looking at experiments, as discussed in Section~\ref{sec:experiments}.


In this section we introduce a smart decomposition algorithm that aims at addressing the above issues; it is designed to be integrated into an ASP system, and uses information available during the computation to predict, according to proper criteria, whether decomposing will pay off or not; moreover, it chooses the most promising decomposition, among the several possible ones.
In the following, we first describe the method in its general form, that can be easily adapted to different real systems; a complete actual implementation, specialized for the \dlv system, is presented later on.

\medskip

The abstract algorithm \decalg is shown in Figure\re{fig:decomp1}; we indicate as \textit{tree decomposition} an actual tree decomposition of a hypergraph, while with \textit{rule decomposition} we denote the conversion of a tree decomposition into a set of ASP rules.
Given a (non-ground) input rule $r$, the algorithm first heuristically computes, by means of the \textsc{Estimate} function, a value $e_r$ that estimates how much the presence of $r$ in the program impacts on the whole computation; then, the function \textsc{GenerateRuleDecompositons} computes a set of possible rule decompositions $RDS$, from which \textsc{ChooseBestDecomposition} selects the best $RD \in RDS$; hence, function \textsc{EstimateDecomposition} computes the value $e_{RD}$ that estimates the
impact of having $RD$ in place of $r$ in the input program. Eventually, function
\textsc{DecompositionIsPreferable} is in charge of comparing $e_r$ and $e_{RD}$ and deciding if decomposing is convenient.
We remark that functions \textsc{Estimate}, \textsc{ChooseBestDecomposition}, \textsc{EstimateDecomposition} and \textsc{DecompositionIsPreferable} are left unimplemented, as they are completely customizable; they must be implemented by defining proper criteria that take into account features and information within the specific evaluation procedure, and the actual ASP system the algorithm is being integrated into.

\begin{figure}
\begin{small}
    \begin{tcolorbox}
    \begin{algorithmic}
        \Function{SmartDecomposition}{$r$ : Rule} : RuleDecomposition
        	\State \textbf{var} $e_r$ : number, $RDS$ : SetOfRuleDecompositions, $e_{RD}$ : Number,
            \State $RD$: RuleDecomposition
        	\State $e_r\gets$ \Call{Estimate}{$r$}
            \State $RDS\gets$ \Call{GenerateRuleDecompositions}{$r$}
            \If{$RDS \not= \emptyset$} \Comment{$r$ is decomposable}
                \State $RD\gets$ \Call{ChooseBestDecomposition}{$RDS$,$e_r$}
            	\State $e_{RD}\gets$ \Call{EstimateDecomposition}{$RD$}
			         \If{\Call{DecompositionIsPreferable}{$e_r$,$e_{RD}$}}
				        \State \textbf{return} $RD$
			         \EndIf
            \EndIf
            \State \textbf{return} $\emptyset$
        \EndFunction
    \end{algorithmic}

    \medskip
        \begin{algorithmic}
        \Function{\textsc{GenerateRuleDecompositons}}{$r$ : Rule} : SetOfRuleDecompositions
        	\State \textbf{var} $HG$ : Hypergraph, $RDS$ : SetOfRuleDecompositions,
            \State $RD$ : RuleDecomposition, $TD$ : TreeDecomposition
            \State $TDS$ : SetOfTreeDecompositions
            \State $HG\gets$ \Call{ToHypergraph}{$r$}
            \State $TDS \gets$ \Call{GenerateTreeDecompositions}{$HG$}
            \For{\textbf{each} {$TD\in TDS$}}
                \State $RD\gets$ \Call{ToRules}{$TD$,$r$}
                \State $RDS = RDS \cup \{RD\}$
          	\EndFor
          	\State \textbf{return} $RDS$
        \EndFunction
    \end{algorithmic}
\caption{\small The algorithm \decalg and the \textsc{GenerateRuleDecompositons}
function.\label{fig:decomp1}}
\end{tcolorbox}
\end{small}
\end{figure}

Figure\re{fig:decomp1} reports also the implementation of function
\textsc{GenerateRuleDecompositons}. Here, \textsc{ToHypergraph} converts a input rule $r$ into a
hypergraph $HG$, which is iteratively analysed in order to produce possible tree decompositions, by
means of the function \textsc{GenerateTreeDecompositions}. Also these stages can be customized in
an actual implementation, according to different criteria and the features of the system at hand;
for space reasons, we refrain from going into details that are not relevant for the description of
the approach. The function \textsc{ToRules}, given a tree decomposition $TD$ and a rule $r$,
converts $TD$ into a rule decomposition $RD$ for $r$. In particular, for each node in $TD$, it adds
a new logic rule to $RD$, possibly along with some additional auxiliary rules needed for ensuring
safety. The process is, again, customizable, and should be defined according to the function
\textsc{ToHypergraph}.

The general definition of the algorithm provided so far is independent from any actual
implementation, and its behaviour can significantly change depending on the customization choices, as discussed above.
However, in order to give an intuition on how it works, we make use of our running example for illustrating a plausible execution.

\begin{example}\label{ex:generalAlgorithm}
Given rule $r_1$ of Example~\ref{ex:nomystery}, let us imagine that function
\textsc{GenerateRuleDecompositions} computes the tree decompositions $TD_1(r_1)$ and
$TD_2(r_1)$ and then, by means of \textsc{ToRules}, the set of rule decompositions consisting of $RD_1(r_1)$ and $RD_2(r_1)$ is generated. Note that $r_4$ and $r_7$ are added for ensuring safety of rules $r_3$ and $r_6$, respectively. Next step consists of the choice between $RD_1(r_1)$ and $RD_2(r_1)$ for the best promising decomposition, according to the actual criteria of choice.
Supposing that it is $RD_1(r_1)$, \textsc{DecompositionIsPreferable} compares the estimated impacts $e_{r_1}$ and $e_{RD_1(r_1)}$ , in order to decide if keeping $r$ or substituting it with
$RD_1(r_1)$.
\end{example}

\section{Integrating the \decalg Algorithm into a Real System: the \dlv Case}\label{subsec:specdecalg}
In this section we illustrate how the general \decalg algorithm of Section~\ref{sec:general} can be customized in order to be integrated into an actual ASP implementation.
Interestingly, such customization can be tailored with different purposes, for both the two-phase-based and the lazy-grounding-based systems, for optimizing solving or instantiation performance, according to different criteria (times, size, structure, etc.).
In this work, we focus on the widespread \dlv system~\cite{DBLP:conf/lpnmr/AlvianoCDFLPRVZ17,DBLP:journals/tocl/LeonePFEGPS06}, which complies to the two-phase strategy, with the explicit aim of optimizing performance of its grounding subsystem \idlv~\cite{DBLP:journals/ia/CalimeriFPZ17}.
A detailed description of the \idlv computation is out of the scope of this work (the interested reader is referred to~\cite{DBLP:journals/ia/CalimeriFPZ17}); however, for the sake of readability, we briefly recall the basics of its machinery.

Given an ASP program $P$:

\begin{enumerate}
\item
    $P$ is parsed, and the extensional database (\EDB) is built.
\item
    Each rule in $P$ is analyzed, and possibly rewritten according to different strategies for optimization purposes; the result constitutes the intensional database (\IDB).
\item
    Dependencies among \IDB\ rules and predicates are examined; such dependencies induce the splitting of $P$ into modules, and a suitable processing ordering is computed so that an incremental evaluation is possible according to the definitions in~\cite{DBLP:conf/birthday/FaberLP12}.
\item\label{item:istanziaterule}
    The program is grounded one module at a time by means of a proper adaptation of a semi-na\"{i}ve schema~\cite{DBLP:conf/birthday/FaberLP12,DBLP:books/cs/Ullman88} that evaluates each rule in a module according to a rule instantiation procedure that in turn produces its ground instances. Rules within a module can be \emph{recursive} or not. While for the former ones the procedure might be iteratively invoked, for the not recursive rules a single call of the rule instantiation procedure is enough to produce all their ground instances.
\item
    The collection of the ground rules generated from all \IDB\ rules compose, along with $EDB(P)$, the resulting ground program $G_P$.
\end{enumerate}




The {\em core} of the \idlv computation is the rule instantiation process mentioned in the step\re{item:istanziaterule} of the sketch above, which constitutes one of the more computationally heavy tasks.
Basically, when grounding a rule $r$ of $P$, instead of replacing $bodyvar(r)$ with every possible constant appearing in $P$, the rule instantiation iteratively substitutes the variables in each body literal with constants appearing in the corresponding predicate extension.
A predicate extension of a predicate $p$ is the set of all ground atoms having $p$ as predicate.
More in detail, given a rule $r$ and the set of extensions of its body predicates, the rule instantiation produces ground instances of $r$ by iterating on positive body literals\footnote{Because of the safety condition, in order to generate a completely ground instance of $r$, it is enough to have a substitution for the variables occurring in the positive literals.} and looking for all possible valid substitutions.
Intuitively, this phase resembles the evaluation of relational joins on the positive body literals, where predicate extensions can be seen as tables whose tuples consist of the ground instances.
Once a valid substitution is found for all variables in $bodyvar(r)$, it is applied to $headvar(r)$ in order to obtain a totally ground rule, i.e. a ground instance of $r$, say $r'$.
This possibly leads to the generation of new ground atoms occurring in the head of $r'$; such new ground atoms are added to the corresponding predicate extensions.
It is worth noting that, the set of all predicate extensions is built dynamically starting from ground atoms appearing in $Facts(P)$ and then, adding each new ground atom coming from heads of produced ground rules; the chosen evaluation order plays a key role in this respect as it ensures that when evaluating a rule $r$ the extensions of all body predicates needed for instantiating $r$ have been fully generated.
%
%

Besides the basic schema herein sketched, \idlv employs smart optimizations techniques,
geared towards the efficient production of a ground program that is considerably  smaller, still preserving the semantics.
Roughly, when a rule is going to be instantiated, \idlv firstly performs a pre-processing that might lead some adjustments over the rule to different extents, and after that the actual rule instantiation takes place, a post-processing refines the output.
Some optimizations, such as, for instance, join-ordering strategies, operate in the pre-processing phase; some explicitly take place during the actual instantiation process, such as non-chronological backtracking; some operate across the two phases, such as indexing techniques for a quick instances retrieval; others take place in the post-processing step, such as the simplification that removes ground rules and literals in the bodies that do not contribute to the semantics.

The \decalg algorithm implementation herein described works in the pre-processing phase.

\medskip

We provide next some details on how we defined the functions that have been left unimplemented in the general description of Section~\ref{sec:general} (\textsc{Estimate},
\textsc{ChooseBestDecomposition}, \textsc{EstimateDecomposition} and
\textsc{DecompositionIsPreferable}), along with the proposed heuristics, and discuss further implementation issues.

\subsection{The \textsc{Estimate} Function}\label{sec:estimation}
The function \textsc{Estimate} (Figure\re{fig:decompIDLV1}) heuristically measures the cost of instantiating a rule $r$ before it is actually grounded.
To this aim, we propose a heuristics inspired by the ones introduced in the database field\cit{DBLP:books/cs/Ullman88} and adopted in\cit{DBLP:conf/lpnmr/LeonePS01} to estimate the size of a join operation. In particular, it relies on
statistics over body predicates, such as size of extensions and argument selectivities; we readapted it in order to estimate the cost of grounding a rule as the total number of operations needed in order to perform the task, rather than estimate the size of the join of its body literals.
Let $a=p(t_1,\ldots,t_n)$ be an atom; we denote with $var(a)$ the set of variables
occurring in $a$, while $T(a)$ represents the number of different tuples for $a$ in the ground extension of $p$. Moreover, for each variable $X \in var(a)$, we denote by $V(X,a)$ the selectivity of $X$ in $a$, i.e., the number of distinct values in the field corresponding to $X$ over the ground extension of $p$.
Given a rule $r$, let $\langle a_1,\ldots,a_m\rangle$ be the ordered list of atoms appearing in $B(r)$, for $m>1$.
Initially, the cost of grounding $r$, denoted by $e_{r}$, is set to $T(a_1)$, then the following formula is iteratively applied up to the last atom in the body in order to obtain the total estimation cost for $r$.
More in detail, let us suppose that we estimated the cost of joining the atoms $\langle a_1,\ldots,a_j\rangle$ for $j\in\{1,\ldots,m\}$, and consequently we want to estimate the cost of joining the next atom $a_{j+1}$; if we denote by $A_j$ the relation obtained by joining all $j$ atoms in $\langle a_1,\ldots,a_j\rangle$, then:

\begin{equation}\label{eq:estimateRule}
e_{A_j\Join a_{j+1}} =  \frac{T(a_{j+1})}{{\displaystyle \prod_{X\in idx(var(A_j)\cap var(a_{j+1}))}V(X,a_{j+1})}} \cdot
\prod_{X\in (var(A_j)\cap var(a_{j+1}))}\frac{V(X,A_j)}{dom(X)}
\end{equation}

\noindent where $dom(X)$ is the maximum selectivity of $X$ computed among the atoms in $B(r)$
containing $X$ as variable, and $idx(var(A_j)\cap var(a_{j+1}))$ is the set of the indexing
arguments of $a_{j+1}$. We note that, at each step, once the atom $a_{j+1}$ has been considered,
$V(X,A_{j+1})$, representing the selectivity of $X$ in the virtual relation obtained at step $j+1$,
has to be estimated in order to be used at next steps:

\begin{equation}\label{eq:estimateSelectivity}
\begin{split}
   V(X,A_{j+1})  = V(X, A_{j}) \cdot \frac{V(X, a_{j+1})}{dom(X)}  & \ \ \ \ \ \ \ \ \ \ \text{if } X \in var(A_{j})  \\
   V(X, A_{j+1}) = V(X, a_{j+1}) & \ \ \ \ \ \ \ \ \ \ \text{otherwise }
\end{split}
\end{equation}

Intuitively, the formula tries to determine the cost of grounding $r$, by estimating the total
number of operations to be performed. In particular, the first factor is intended to estimate how
many instances for $a_{j+1}$ have to be considered, while the second factor represents the
reduction in the search space implied by $a_{j+1}$. To obtain a realistic estimate, the presence of
indexing techniques, used in \idlv to reduce the number of such operations\cit{DBLP:journals/ia/CalimeriFPZ17}, has been taken into account.

\begin{example}\label{ex:formula}
Let us consider the rule:
$$r_1: p(X,Y,Z,S)\derives s(S),\ a(X,Y,S-1),\ c(D,Y,Z),\ f(X,P,S-1),\ P >= D.$$
of Example~\ref{ex:nomystery}, and let us assume that we are dealing with an instance that contains the facts\footnote{According
to \aspcore syntax, the term $(1..k)$ stands for all values from $1$ to $k$.}:
$$ s(1..5).\ \ \ a(1..5, 1..5, 1..5).\ \ \ c(1..5, 1..5, 1..5).\ \ \ f(1..5, 1..5, 1..5). $$
The \textsc{Estimate} function first estimates, by means of Formula~(\ref{eq:estimateRule}), the cost of computing the joins $A_i$. In this case, denoting by $a_1 = s(S)$, $a_2 = a(X,Y,S-1)$, $a_3 = c(D,Y,Z)$ and so on, we have that
$A_1 = s(S)$,  $A_2 =$ $A_1 \Join$ $a(X,Y,S-1)$ and it is estimated as:

\begin{equation*}
e_{A_1\Join a_2} =  \frac{T(a_2)}{{\displaystyle V(S,a_2)}} \cdot
\frac{V(S,A_1)}{dom(S)} =  \frac{125}{{\displaystyle 5}} \cdot
\frac{5}{5} = 25 = e_{A_2}
\end{equation*}

Then, the formula is used again in order to estimate the cost of the join $A_3$ between $A_2$ and $a_3$, and so on up to the last join $A_4$.
At each step, size and variable selectivities for each $a_i$ are known, while such data for the intermediate relations $A_i$ are estimated. The size of $A_2$ is estimated as $e_{A_2}$, and selectivities of all variables appearing in $A_2$ (i.e., $X$,$Y$, and $S$) are estimated, according to Formula~(\ref{eq:estimateSelectivity}), as:

\begin{compactitem}
\item[-]
    $V(X,A_2) = 5$ \ \ \ (indeed, $X \notin var(A_1)$)
\item[-]
    $V(Y,A_2) = 5$ \ \ \ (indeed, $Y \notin var(A_1)$)
\item[-]
    $V(S,A_2) = V(S,A_1) \cdot \frac{V(S,a_2)}{dom(S)} = 5$ \ \ \ (indeed, $S \in var(A_1)$).
\end{compactitem}

The process is similarly iterated until the end of the body, from left to right.
\end{example}

\begin{figure}
\begin{small}
\begin{tcolorbox}
     \begin{algorithmic}
        \Function{Estimate}{$r$ : Rule} : Number
            \LineComment{Estimate the cost of grounding a rule according to Formula~(\ref{eq:estimateRule})}
        \EndFunction
    \end{algorithmic}

    \begin{algorithmic}
        \Function{EstimateDecomposition}{$RD$ : SetOfRules} : Number
        \State \textbf{var} $e_{RD}$ : number
        	\State \Call{PreProcess}{$RD$}
        	\State $e_{RD}\gets 0$
			     \For {\textbf{each} $r' \in RD$}
				    \State $e_{RD} = e_{RD} + $ \Call{Estimate}{$r'$}
			     \EndFor
        \State \textbf{return} $e_{RD}$
        \EndFunction
    \end{algorithmic}
\caption{\small  \textsc{Estimate} and \textsc{EstimateDecomposition} as implemented in \idlv}
    \label{fig:decompIDLV1}
\end{tcolorbox}
\end{small}
\end{figure}

\subsection{The \textsc{EstimateDecomposition} Function}\label{sec:estimationDecomp}
The \textsc{EstimateDecomposition} function is illustrated in Figure\re{fig:decompIDLV1}: after
some pre-processing steps, it computes the cost of a given decomposition as the sum of the cost of
each rule in it.
Let $r$ be a rule and $RD=\{r_1,\dots,r_n\}$ be a rule decomposition for $r$. In
order to estimate the cost of grounding $RD$, one must estimate the cost of grounding all rules in
$RD$.
For each $r_i \in RD$ the estimate is performed by means of Formula~(\ref{eq:estimateRule}).
Nevertheless, it is worth noting that each $r_i$, in addition to predicates originally appearing in
$r$, denoted as \textit{known predicates}, may contain some \textit{fresh predicates}, generated
during the decomposition.
Concerning known predicates, thanks to the rule instantiation ordering followed by \idlv, as already pointed out in Section\re{subsec:specdecalg}, extensions size and selectivity needed for computing the formula are directly available: hence, there is no need for estimations.
On the contrary, for fresh predicates, that have been ``locally'' introduced and do not appear in any of the rules in the original input program, such data is not available, and must be estimated.
To this aim, the dependencies among the rules in $RD$ are
analyzed, and an ordering that guarantees a correct instantiation is determined. Such dependencies come out from the definitions in~\cite{DBLP:conf/birthday/FaberLP12}: rules depending only on known predicates can be grounded first, while rules depending also on new predicates can be grounded only once the rules that define them have been instantiated.
Assuming that for the set $RD$ a correct instantiation order is represented by $\langle r_1,\dots,r_n
\rangle$, for each $r'$ in this ordered list, if $H(r')=p'(t_1,\ldots,t_k)$ for $k\geq 1$, and if
$p'$ is a fresh predicate, we estimate: $(i)$ the size of the ground extension of $p'$, denoted $T(p')$,
by means of a formula conceived for estimating the size of a join relation, based on criteria that
are well-established in the database field and reported in~\cite{DBLP:conf/lpnmr/LeonePS01}; $(ii)$ the
selectivity of each argument as $\sqrt[k]{T(p')}$.
Therefore, the procedure \textsc{PreProcess} invoked in \textsc{EstimateDecomposition} (see
Figure~\ref{fig:decompIDLV1}) amounts to preprocess the rules in $RD$ according to a valid
grounding order $\langle r_1,\ldots,r_n \rangle$ to obtain the extension sizes and the argument
selectivities for involved fresh predicates, based on the above mentioned formula.
Once estimates for fresh predicates are available, the actual estimate of grounding $RD$ can be performed.

\begin{example}\label{ex:estimatedecomp}
Let us consider again the rule $r_1$ of our running Example\re{ex:nomystery} and its decomposition $RD_1(r_1)$:
\begin{dlvcode}
    r_2: p(X,Y,Z,S) \derives s(S), a(X,Y,S-1), f(X,P,S-1), fresh\_pred\_1(P,Y,Z).\\
    r_3: fresh\_pred\_1(P,Y,Z) \derives c(D,Y,Z), P>=D, fresh\_pred\_2(P).\\
    r_4: fresh\_pred\_2(P) \derives s(S), f(\anonym,P,S-1).
\end{dlvcode}

In order to compute $e_{RD_1(r_1)}$ we first need to determine a correct evaluation order of the rules in $RD_1(r_1)$; the only valid one is $\langle r_4,r_3,r_2 \rangle$.
Indeed, $r_4$ has only known predicates in its body, thus can be evaluated first; the body of $r_3$ contains, besides to known predicates, $fresh\_pred\_2$, whose estimates will be available just after the evaluation of $r_4$; eventually, $r_2$ depends also on $fresh\_pred\_1$, whose estimates will be available right after the evaluation of $r_3$.
Once the estimates for the fresh predicates $fresh\_pred\_1$ and $fresh\_pred\_2$ are obtained, they are
used for computing $e_{r_4}$, $e_{r_3}$ and $e_{r_2}$ with Formula~(\ref{eq:estimateRule}), and
then for obtaining $e_{RD_1(r_1)}$ = $e_{r_2} + e_{r_3} + e_{r_4}$.
\end{example}

\subsection{The \textsc{ChooseBestDecomposition} and \textsc{DecompositionIsPreferable} Functions}\label{sec:choose}
The function \textsc{ChooseBestDecomposition} 
estimates the costs of all decompositions of a rule $r$ by means of \textsc{EstimateDecomposition}, and returns the one with the smallest estimated cost; let us denote it by $RD$.
The function \textsc{DecompositionIsPreferable} is then in charge of deciding whether $RD$ will substitute $r$, by relying on $e_r$ and $e_{RD}$, that are the estimated costs associated to $r$ and $RD$, respectively.
More in detail, it computes the ratio\ \ $e_r/e_{RD}$.\
Intuitively, when the ratio $e_r/e_{RD} \geq 1$,\ \ decomposing $r$ is convenient; 
nevertheless, it is worth remembering that the costs are estimated, and, in particular, as discussed in Section~\ref{sec:estimationDecomp}, the estimate of the cost of a decomposition requires to estimate also the extension of some additional predicates introduced by the rewriting, thus possibly making the estimate less accurate.
This leads sometimes to cases in which the decomposition is preferable even when $e_r/e_{RD} < 1$.
One can try to improve the estimations, in the first place; however, an error margin will always be present.
For this reason, in order to reduce the impact of such issue, we decided to experimentally test the effects of the choices under
several values of the ratio, and found that decomposition is preferable when $e_r/e_{RD} \geq 0.5$,
that has also been set as a default threshold in our implementation; of course, the user can play
with this at will. We plan to further improve the choice of the threshold by taking advantage from
automatic and more advanced methods, such as machine learning guided techniques.

\begin{example}
Let us consider again our running Example\re{ex:nomystery}.
At the final step, three possible alternatives are evaluated: $(i)$ leave the rule $r_1$ as it is (i.e. $r_1$ is not decomposed), $(ii)$ choose $RD_1(r_1)$ or $(iii)$ choose $RD_2(r_1)$.
Since the nature of the heuristics we implemented into \idlv have the aim of optimizing the grounding process, estimations tightly depend on the instance at hand; hence, choices will possibly vary from
instance to instance.

Let us assume that the current instance contains the same facts reported in Example~\ref{ex:formula}.
Then, the costs of instantiating $RD_1(r_1)$ or $RD_2(r_1)$ are computed according to what discussed in Section~\ref{sec:estimationDecomp}: without reporting all intermediate calculations, we have $e_{RD_1(r_1)}$ $=$ $122,945$, while $e_{RD_2(r_1)}$ $=$ $53,075$.
In this case, the best decomposition is obviously $RD_2(r_1)$, and it is compared with the option of grounding $r_1$ as non-decomposed.
Again, without reporting all intermediate calculations, we have that the cost $e_{r_1}$ of grounding $r_1$ amounts to $390,625$; hence, the ratio $e_{r_1}/e_{RD_2(r_1)}$ is computed as $7.36$, and, given that it is greater than $0.5$, we prefer to substitute the original rule with the decomposition $RD_2(r_1)$ (see discussion above).

Interestingly, with a different input instance, things might change. For instance, if the set of input facts for $f$ changed to $f(1..20, 1..20, 1..5).$, the decomposition $RD_1(r_1)$ would be preferred.
\end{example}

\subsection{Fine-Tuning and Further Implementation Issues}\label{sec:idlv}
In order to implement the \decalg algorithm, one might rely on \lpopt in order to obtain a rule decomposition for each rule in the program; in particular, this would lead to a straightforward implementation of \textsc{ToHypergraph} and \textsc{ToRules}, the functions that convert a rule into a hypergraph and a tree decomposition into a rule decomposition, respectively.
Nevertheless, in order to better take advantage from the features of \idlv and do not interfere with its existing optimizations, we designed ad-hoc versions for such functions.

For instance, \idlv supports the whole \aspcore language, which contains advanced constructs like
aggregates, choice rules and queries; our implementation, even if resembling the one of \lpopt,
introduces custom extensions explicitly tailored to \idlv optimizations, and some updates in the
way the aforementioned linguistic extensions are handled.
It is worth noting that, when dealing with rules containing aggregate literals or choice atoms~\cite{asp-core-2-01c}, \idlv
rewrites them: briefly, each conjunction of literals in aggregate and/or choice elements is replaced by a fresh atom, and an auxiliary rule is added to preserve semantics; this ensures more efficiency and transparency
with respect to \idlv grounding machinery and its native optimization techniques.
As a result, the \decalg algorithm, that takes place after such rewritings, does not need to explicitly take care of internal conjunctions of aggregates or choice constructs; on the contrary, \lpopt possibly decomposes also such internal conjunctions.

Differently from \lpopt, \idlv explicitly handles queries, and employs the magic sets rewriting technique\cit{DBLP:journals/ai/AlvianoFGL12} to boost query answering; 
in our approach, \decalg is applied after the magic rewriting has occurred, so that decompositions
is applied also to resulting magic rules. In addition, given that \idlv performs other rewritings
on the input rules for optimization purposes, the function \textsc{ToRules} is in charge of
performing such already existing rewriting tasks also on the rules resulting from the
decompositions.

Another relevant issue is related to the safety of the rules generated in a decomposition. Indeed,
due to the abstract nature of \decalg, we cannot assume that they are safe, since this depends on
the schemas selected for converting a rule into a hypergraph, and a tree decomposition into a set
of rules. Hence, the \textsc{ToRules} function must properly take this into account, as briefly
noted in Section~\ref{sec:general}. In particular, our implementation, given a rule $r$ and an
associated tree decomposition $TD$, after a rule $r'$ corresponding to a node in $TD$ has been
generated, checks its safety. If $r'$ is unsafe, and $UV$ is the set of unsafe variables in $r'$,
an atom $a$ over a fresh predicate $p$, that contains the variables in $UV$ as terms, is added to
$B(r')$ and a new rule $r''$ is generated, having $a$ as head; a set of literals $L$ binding the
variables in $UV$ is extracted from $B(r)$ and added to $B(r'')$. Interestingly, the choice of the
literals to be inserted in $L$ is in general not unique, as different combinations of literals
might bind the same set of variables; for instance, one might even directly add $L$ to $B(r')$
without generating $r''$; however, this might introduce further variables in $B(r')$, and alter the
original join operations in it. For this reason, in our implementation we decided to still add
$r''$, and while choosing a possible binding, for each variable $V\in UV$ we try to keep the number of literals taken from $B(r)$ small, also preferring to pick positive literals with small ground extensions.

More in detail, given a variable $V\in UV$, we look for a ``standard'' positive literal $l$ that binds $V$ and features as terms only variables, constants or functional terms: the rationale behind such choice is that no additional literals will be needed to guarantee the safety of $l$ itself.
If more than one such literals exists, we select the one with the smallest extension size; if no one is available, we pick up the first suitable literal according to the following predefined priority order: classical literals featuring other kind of terms (such as arithmetic terms), built-in atoms and aggregate literals, respectively. Intuitively, for such literals, additional ones from $B(r)$ may in turn be needed to ensure their safety. Interestingly, the choice of saviour literals is more careful than what it would be obtained by using \lpopt as a black box, as in this case the choice could not rely on information that are available only from within the instantiation process.

\medskip

The current implementation of function \textsc{GenerateTreeDecompositions}, which, given a hypergraph $HG$, is in charge of returning a set of tree decompositions $TDS$, relies on the open-source \texttt{C++} library \texttt{htd}\cit{DBLP:conf/cpaior/AbseherMW17}\footnote{\url{https://github.com/mabseher/htd}}, an efficient and flexible library for computing customized tree and hypertree decompositions; in our implementation, we used the most recent version available at the time of writing.
The library features several methods for computing tree decompositions according to different heuristics described in literature; we took advantage from this, and our implementation allows the user to deviate from the default method via a command-line option.
Interestingly, the \texttt{htd} library features also a fitness mechanism for ``ranking'' decompositions according to a user-provided fitness function.
In our setting, we made use of such mechanism in order to associate a cost estimation relying on Formula~(\ref{eq:estimateRule}) (see Section~\ref{sec:estimation}) to a computed decomposition; hence, in the decomposition selection phase, \idlv generates a number of tree decompositions and selects the best one as the one the lowest instantiation cost, according to our criteria.
This constitutes another important difference w.r.t. the approach of \lpopt, that instead, makes use of the same generation tool in order to obtain just one decomposition per each rule with no evaluation at all.
Obviously, handling the fitness mechanism can imply some overhead w.r.t. the choice of computing only one, unevaluated, decomposition.
For this reason, in our approach decompositions are requested and evaluated one at a time and, in order to limit the impact of such phase on performance, \idlv by default stops the generation of additional tree decompositions after $3$ consecutive generations that do not show improvements in the fitness values, and never looks for more than a total of $5$ generations.
These limits have been set by experimentally observing that the consecutive decompositions generated by \texttt{htd} present no performance improvements with higher values; however, they can be customized by means of command-line options.
%
%
The selected decomposition is then compared against the original rule in order to check whether it is convenient to actually decompose or not, as described in Section\re{sec:estimation}.

An additional expedient to limit overheads consists in disabling the fitness mechanism in case of rules featuring very long bodies; indeed, computing multiple decompositions may be particularly costly for such rules (see Section~\ref{sec:experiments}).
Therefore, we set a limit to body length so that, when it is exceeded, the fitness mechanism is automatically disabled and just one decomposition is generated and checked against the original rule.
Again, we set a default value experimentally; by default, the limit is set to $10$ literals per body, but it can be changed by means of a command-line option.
In this respect, a possible improvement of our technique, which will be subject of future work, consists in properly generating a unique, presumably ``good'' enough, decomposition, thus preventing the expensive production of multiple ones.

\section{Experimental Evaluation}\label{sec:experiments}

We carried out a thorough experimental activity aimed at assessing the impact of \decalg on the grounding performance of \idlv, analyzing the effectiveness of the proposed heuristics, and also at having a first glance on the effect of the produced instantiation over state-of-the-art ASP solvers.
For the sake of readability, we discuss next only a significant subset of the experiments that have been carried out; additional experiments are illustrated in appendices.
%

\subsection{Benchmarks and Results}\label{subsec:results}

All the experiments reported in this section and in the following have been performed on a NUMA machine equipped with two {\small\texttt{2.8 GHz AMD Opteron 6320}} and {\small\texttt{128 GiB}} of main memory, running {\small\texttt{Linux Ubuntu 14.04.4 (kernel ver. 3.19.0-25)}}. Binaries have been generated by the {\small\texttt{GNU C++}} {\small\texttt{compiler}} {\small\texttt{5.4.0}}.
We allotted {\small\texttt{15 GiB}} and {\small\texttt{600}} seconds to each system per each single run, as memory and time limits.
Three versions of \idlv have been compared: $(i)$ \idlv without any decomposition, $(ii)$ \lpopt (version {\small\texttt{2.2}}) combined in pipeline with \idlv (i.e., a black-box usage of \lpopt), $(iii)$ \idlvsd, i.e. \idlv empowered with the herein introduced version of \decalg.

As for benchmarks, we first considered the whole $6th$ ASP Competition suite~\cite{DBLP:conf/lpnmr/GebserMR15}, the latest available at the time of writing;
for each problem, the average time over the $20$ selected instances of the official Competition runs is reported; in order to produce replicable results, the random seed used by \lpopt for heuristics has been set to $0$ for system $(ii)$.

\begin{table}
  \caption{\small $6th$ Competition (20 instances per problem) -- Grounding Benchmarks: number of grounded instances and average running times (in seconds). \texttt{US} indicates that corresponding configurations do not support the adopted syntax.\label{table:grounding}}
  \centering
  \tabcolsep=0.060cm
    \begin{tabular}{lrrrrrrrr}
    \hline\hline
    \multirow{2}[2]{*}{\textbf{Problem}} & \multicolumn{2}{c}{\textbf{\idlv}} & \multicolumn{2}{c}{\textbf{\lpopt $\vert$ \idlv}} & \multicolumn{2}{c}{\textbf{\idlvsd}} & \multicolumn{2}{c}{\textbf{\idlvsd gap}} \\
    \cmidrule{2-3}     \cmidrule{4-5}      \cmidrule{6-7}   \cmidrule{8-9}
    & \textbf{\#grnd} & \textbf{time} & \textbf{\#grnd} & \textbf{time} & \textbf{\#grnd} & \textbf{time} & \textbf{absolute} & \textbf{\%} \\
    \hline
Abstract Dialectical Frameworks & 20      & 0,12   & 20            & 0,12   & 20      & 0,12   & 0,00                    & 0\%                          \\
Combined Configuration          & 20      & 13,58  & 20            & 13,39  & 20      & 13,15  & 0,24                    & 2\%                          \\
Complex Optimization            & 20      & 57,56  & 20            & 60,72  & 20      & 57,24  & 0,32                    & 1\%                          \\
Connected Still Life            & 20      & 0,10   & 20            & 0,10   & 20      & 0,10   & 0,00                    & 0\%                          \\
Consistent Query Answering      & 20      & 76,44  & 0             & US     & 20      & 77,00  & -0,57                   & -1\%                         \\
Crossing Minimization           & 20      & 0,10   & 20            & 0,10   & 20      & 0,10   & 0,00                    & 0\%                          \\
Graceful Graphs                 & 20      & 0,30   & 20            & 0,31   & 20      & 0,30   & 0,00                    & 0\%                          \\
Graph Coloring                  & 20      & 0,10   & 20            & 0,10   & 20      & 0,10   & 0,00                    & 0\%                          \\
Incremental Scheduling          & 20      & 16,07  & 20            & 15,74  & 20      & 16,21  & -0,47                   & -3\%                         \\
Knight Tour With Holes          & 20      & 1,83   & 20            & 5,98   & 20      & 1,84   & -0,01                   & -1\%                         \\
Labyrinth                       & 20      & 1,97   & 20            & 1,83   & 20      & 2,02   & -0,18                   & -10\%                        \\
Maximal Clique                  & 20      & 4,93   & 20            & 21,60  & 20      & 4,96   & -0,03                   & -1\%                         \\
MaxSAT                          & 20      & 3,85   & 20            & 8,87   & 20      & 3,86   & -0,01                   & 0\%                          \\
Minimal Diagnosis               & 20      & 5,09   & 20            & 4,30   & 20      & 4,22   & 0,07                    & 2\%                          \\
Nomistery                       & 20      & 3,45   & 20            & 1,94   & 20      & 3,63   & -1,68                   & -87\%                        \\
Partner Units                   & 20      & 0,46   & 20            & 0,47   & 20      & 0,47   & 0,00                    & 0\%                          \\
Permutation Pattern Matching    & 20      & 130,47 & 20            & 4,35   & 20      & 4,21   & 0,14                    & 3\%                          \\
Qualitative Spatial Reasoning   & 20      & 5,44   & 20            & 5,50   & 20      & 5,44   & 0,00                    & 0\%                          \\
Reachability                    & 20      & 126,54 & 0             & US     & 20      & 126,14 & 0,40                    & 0\%                          \\
Ricochet Robots                 & 20      & 0,36   & 20            & 0,39   & 20      & 0,39   & -0,03                   & -9\%                         \\
Sokoban                         & 20      & 1,21   & 20            & 1,23   & 20      & 1,22   & -0,01                   & -1\%                         \\
Stable Marriage                 & 20      & 118,55 & 20            & 125,78 & 20      & 119,53 & -0,99                   & -1\%                         \\
Steiner Tree                    & 20      & 29,00  & 20            & 28,92  & 20      & 29,11  & -0,19                   & -1\%                         \\
Strategic Companies             & 20      & 0,19   & 0             & US     & 20      & 0,20   & 0,00                    & -1\%                         \\
System Synthesis                & 20      & 1,09   & 20            & 1,15   & 20      & 1,08   & 0,01                    & 1\%                          \\
Valves Location Problem         & 20      & 2,52   & 20            & 2,53   & 20      & 2,54   & -0,02                   & -1\%                         \\
Video Streaming                 & 20      & 0,10   & 20            & 0,10   & 20      & 0,10   & 0,00                    & 0\%                          \\
Visit-all                       & 20      & 1,18   & 20            & 0,44   & 20      & 0,48   & -0,04                   & -9\%                         \\
Total Grounded Instances & \multicolumn{2}{c}{560/560} & \multicolumn{2}{c}{500/560} & \multicolumn{2}{c}{560/560} & & \\
\hline\hline
\end{tabular}%
\end{table}

\begin{figure}
  \centering
  \includegraphics[width=\textwidth]{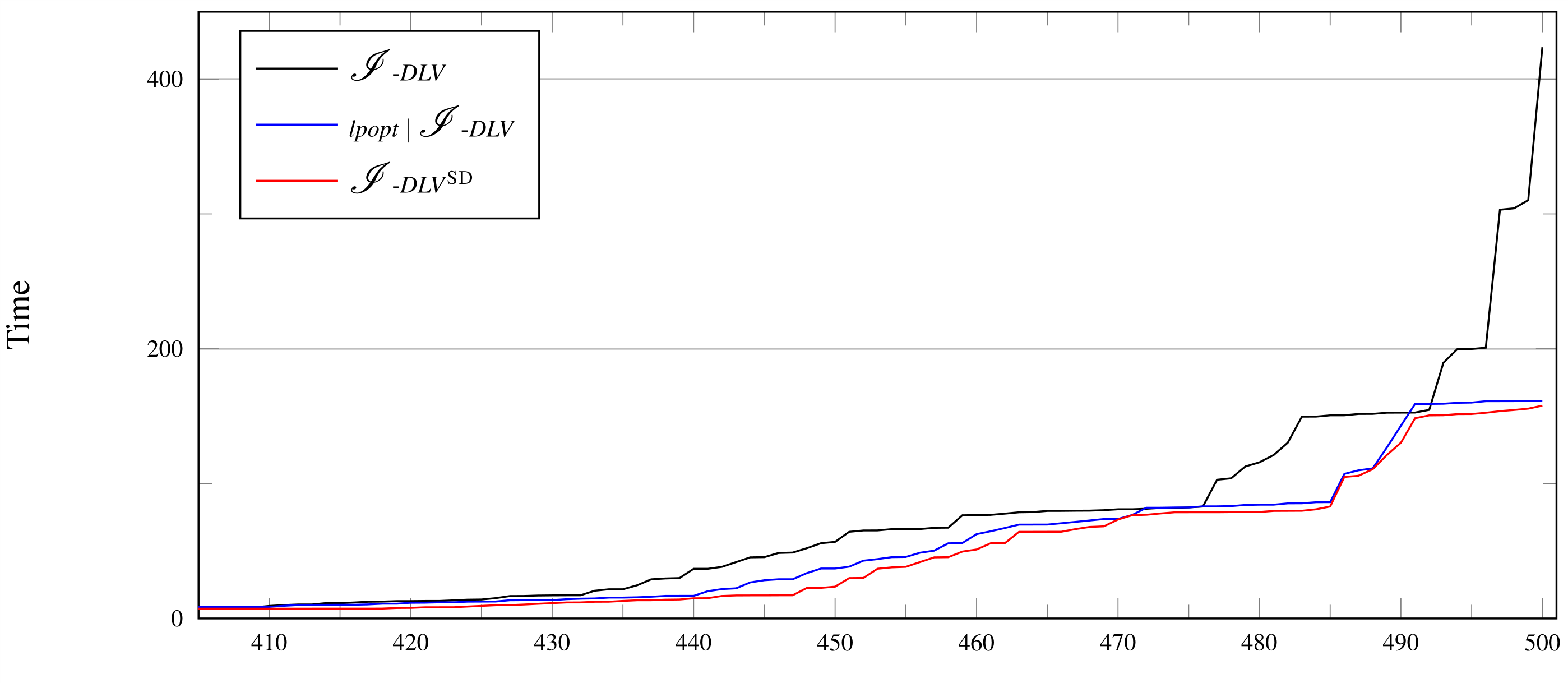}
        \vspace{1ex}
        \caption{6th Competition Grounding Benchmarks (excluding domains featuring queries): grounded instances over time (in seconds).\label{fig:6thCompGroudingbench1}}
\end{figure}

Results are reported in Table~\ref{table:grounding}, showing number of grounded instances within the allotted time along with the average time spent.
The symbol \verb"US" in the table indicates that a configuration does not support the syntax of the encoding for corresponding domain; in particular, this happens in case of domains featuring queries, as system $(ii)$ is not able to process them because of the lack of support for queries from \lpopt.


Results of the ``blind usage'' of \lpopt (system $(ii)$) are conflicting: for instance, in some cases it enjoys a great gain w.r.t. the version of \idlv without decomposition, in particular while dealing with the \textit{Permutation Pattern Matching} problem, yet showing great losses in other cases, such as \textit{Knight Tour With Holes}, where instantiating rules resulting from the decomposition requires more time w.r.t. the input ones.
On the other hand, from Table~\ref{table:grounding} it is easy to see that the \decalg algorithm allows \idlvsd to always match or overcome \idlv performances, still enjoying relevant improvements when decomposition is actually convenient (up to $96.7$\% in case of \textit{Permutation Pattern Matching}), and avoiding negative effects of the black-box decomposition mechanism, as in the case of \textit{Knight Tour With Holes}.
In addition, we note also that the \idlvsd is able to limit the overhead w.r.t. \idlv: indeed, it is negligible even in cases where decomposition does not pay; the same does not hold for the system $(ii)$ which suffers from the useless additional invocation of \lpopt in all cases when the input program cannot be decomposed (see, e.g., \textit{Maximal Clique}).

As a remark, what we expected here is that, while dealing with such benchmarks whose encodings coming from the ASP competition are already highly optimized, \idlvsd performed similarly to \idlv (with no decompositions) in all cases where decomposition is not convenient, and similarly to system $(ii)$ otherwise.
In order to assess this, we computed absolute and relative differences in term of times between \idlvsd and the best performing among the other two configurations, for each benchmark; data are reported in the two rightmost columns of Table~\ref{table:grounding}.
As it can be observed, apart from negligible fluctuations and with the only relevant exception consisting of \emph{Nomystery}, our expectations have been met: absolute differences are close to zero, meaning that \idlvsd behaviour is systematically comparable with the best one among the other two.
The special case of \emph{Nomystery} is discussed later in this section.

An additional view of the general picture coming from this set of benchmarks is given by the plot in Figure\re{fig:6thCompGroudingbench1}, built over the same data of Table~\ref{table:grounding} except for the three domains featuring queries, that, as already mentioned, are unsupported by system $(ii)$.
The plot allows us to appreciate the advantage granted by the decomposition rewriting, as both systems $(ii)$ and $(iii)$ clearly outperform system $(i)$, and to note that the performance reached thanks to the \decalg algorithm are consistently better than what achieved via the unconditional use of decomposition.

\medskip

\begin{table}
  \caption{2QBF Grounding Benchmarks: number of total grounded instances.\label{table:qbfbench}}
  \centering
  \resizebox{\columnwidth}{!}{%
    \begin{tabular}{ccc}
    \hline\hline
    \textit{\textbf{\idlv}} & \textit{\textbf{\lpopt}} $\vert$ \textit{\textbf{\idlv}} & \textit{\textbf{\idlvsd}} \\
    \hline
    8 & 82 & 96 \\
    \hline\hline
    \end{tabular}%
    }
\end{table}

\begin{figure}
  \centering
  \includegraphics[width=\textwidth]{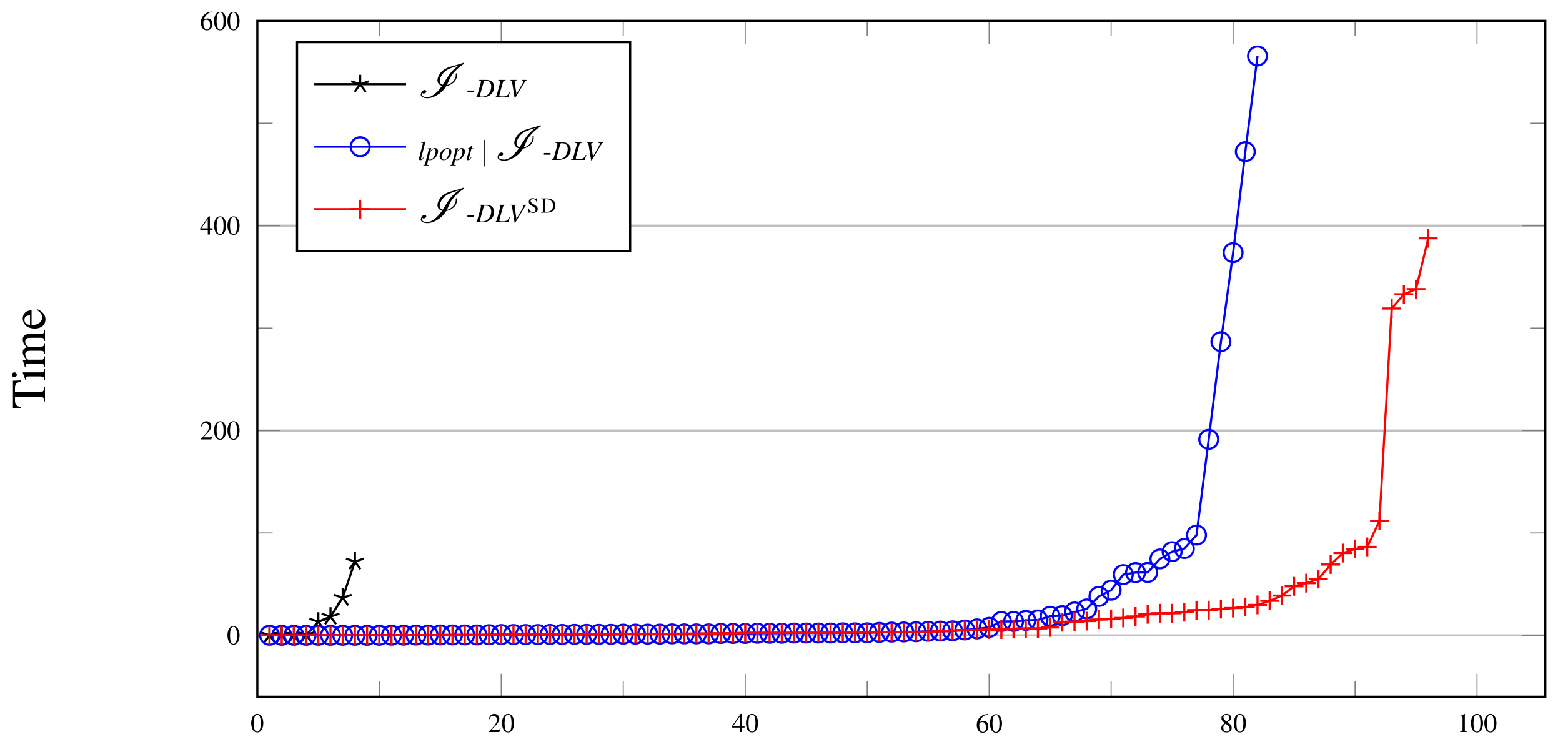}
  \caption{2QBF Grounding Benchmarks: grounded instances over time (in seconds).\label{fig:qbfbench1}}
\end{figure}
%
%
%
%

Furthermore, we considered an additional set of benchmarks that
have been already used in\cit{DBLP:journals/tplp/BichlerMW16} in order to test the efficiency
of ASP-solvers paired with \lpopt over challenging programs.
In particular, in\cit{DBLP:journals/tplp/BichlerMW16}, some publicly available QBF instances have been ported to ASP, according to a conversion strategy that
produces programs featuring a complex structure and very long rules.
%
This test-suite includes $200$ ASP programs, each one corresponding to a different 2-QBF instance.
The results are depicted in Figure\re{fig:qbfbench1}: the number of grounded instances is on the x-axis while running times (in seconds) are on the y-axis; the total number of successfully grounded instances per each tested configuration is reported in Table\re{table:qbfbench}.
%
First of all, we note that while dealing with these problems applying a decomposition on decomposable rules is always a good choice; indeed, when no decomposition is performed the number of grounded instances is significantly smaller and running times are higher w.r.t. configurations adopting decomposition techniques. Moreover, the heuristics guiding \decalg in \idlv work properly, estimating the decompositions as convenient, and \idlvsd automatically disables, internally, the fitness mechanism because the body size of rules is higher than the fixed limit, thus reducing the risk of high overheads (cf. Section\re{sec:idlv}).
In general, \lpopt$\vert$\idlv and \idlvsd enjoy similar performance, and \idlvsd behaves as the best performing version.
Although both versions decompose the same rules, \idlvsd benefits from a tight integration of the decomposition mechanism into the evaluation process that allows to better interact with the other optimization strategies of \idlv and possibly lead to different choices in decompositions.
Finally, on the technical side, we note that, \lpopt and \idlvsd rely on different versions of the \texttt{htd} library versions.


\medskip

In summary, results of experiments clearly show the effectiveness of the herein proposed approach.
Some further considerations can be done about implementation and integration into a system like \idlv.
Besides merely technical aspects, it is worth remembering that, as already mentioned, \idlv is packed with a large number of optimizations; this means that a rewriting-based technique such as the \decalg algorithm might have non-trivial interactions with them.
Our experiments show that, in general, these interactions lead to performance gains, as it is clear while looking, for instance, at the QBF problem; nevertheless, a few isolated   cases go towards different outcomes.
In particular, looking at Table~\ref{table:grounding}, we find that \emph{Nomystery} and, even if to a smaller extent, \emph{Labyrinth}, apparently benefit more from the black-box usage than from the heuristic-guided one.
However, this is not the case: we investigated, and found that the reason is not related to the choices made according to the heuristics, but rather to the mentioned interaction with other internal rewritings performed by \idlv before the decomposition stage (for more details, we refer the reader to~\cite{DBLP:journals/ia/CalimeriFPZ17}); a more detailed study of such interaction will be subject of future works.

\subsection{On the Effectiveness of the Heuristics}\label{subsec:effectiveness}
In order to better understand the actual effects on grounding performance of the \decalg algorithm as guided by the proposed heuristics, we computed some relevant statistics starting from the data obtained from experiments over all domains considered in our experimental activities, thus including, besides those described in Section~\ref{subsec:results}, those described in Appendices; we aggregate them over specific set of instances, as described next, and report the results in a table comparing the behaviours of \idlv and \idlvsd.
In particular, Table~\ref{table:impact-1} shows two sets of data: the first refers to the whole collection of problem domains, while the second to the subset of ``affected domains'', i.e., problems where significant differences on performance are reported, either positive or negative\footnote{A problem domain is here considered as ``affected'' if either the number of instances grounded by the two systems differ, or, in case the number is the same, difference in average grounding times between the two systems is either above $+10\%$ or below $-10\%$.}.
The first two columns report performance of the two system configurations, respectively, while the third reports the percentage gain achieved by \idlvsd thanks to the \decalg algorithm.

\begin{table}
\caption{Detailed comparison of \idlvsd against \idlv.\label{table:impact-1}}
  \centering
  \tabcolsep=0.25cm
    \begin{tabular}{lrrrrrr}
    \hline\hline
                                                        &                     & \idlv & \idlvsd & \idlvsd gain \\ \midrule
\multicolumn{1}{r|}{\multirow{4}{*}{All problems}}      & \#solved instances  & 1688  & 1787    & 6\%          \\
\multicolumn{1}{r|}{}                                   & Average time        & 22,26 & 15,39   & 31\%         \\
\multicolumn{1}{r|}{}                                   & \# timeouts         & 113   & 19      & 83\%         \\
\multicolumn{1}{r|}{}                                   & \# memouts          & 5     & 0       & 100\%        \\ \midrule
\multicolumn{1}{l|}{\multirow{4}{*}{Affected problems}} & \# solved instances & 392   & 491     & 25\%         \\
\multicolumn{1}{l|}{}                                   & Average time        & 40,10 & 10,28   & 74\%         \\
\multicolumn{1}{l|}{}                                   & \# timeouts         & 103   & 9       & 91\%         \\
\multicolumn{1}{l|}{}                                   & \# memouts          & 5     & 0       & 100\%        \\ 
    \hline\hline
    \end{tabular}%
\end{table}

It is easy to see that the positive impact of the technique on grounding performance, on the overall (i.e., over all problems), is significant: a hundred of additional grounded instances ($+6$\%), more than $80$\% of timeouts avoided, and no more instances remain unsolved because of the excessive amount of required memory.
The impact is even more evident if we consider that average times are computed only over the set of instances that are solved by both \idlvsd and \idlv; still, the performance gain turns out to be over $30$\%.

When we focus on the set of affected problems, the benefits of the proposed techniques are even more evident; we just note here as the gain in average grounding times, still computed only over the set of instances that are grounded by both systems, is almost $75$\%.

\subsection{Impact of \idlv\!\!$^{SD}$ on ASP Solvers}\label{exp:solving}
We proved above how a smart decomposition strategy significantly improves performance of a grounder like \idlv; interestingly, such improvements on the instantiation process are relevant from many perspectives.
First of all, as already mentioned in the introduction, a grounder like \idlv is actually a full-fledged deductive database system, that can profitably employed in many real-world domains for non-trivially querying knowledge bases of various nature, ranging from traditional relational to ontology-based ones. In these contexts, typically, programs to be evaluated turn out to be normal and stratified, and thus, completely solvable by a proper grounder. In all such cases, given that solving phase is not needed, each improvement on the grounding side trivially implies improvements on the whole ASP computation.
In addition, the proposed technique can be of great help in all those cases where, given the nature of standard ASP evaluation strategy, the ground program can be so huge that it constitutes a bottleneck. The \decalg algorithm can be useful for mitigating this issue, allowing to actually instantiate programs that cannot be grounded without: let us think, for instance, of the 2-QBF domain discussed in Section~\ref{subsec:results}.
Furthermore, even if the proposed technique aims at improving grounding, it has a positive impact also on solving times, thus allowing to improve performance of the whole computational process.
To evaluate such impact we performed an additional experimental analysis; in particular, we combined the same three versions of \idlv used in Section~\ref{subsec:results} with the two mainstream ASP solvers \clasp~\cite{DBLP:conf/lpnmr/GebserKK0S15} (version {\small\texttt{5.2.1}}) and \wasp~\cite{DBLP:conf/lpnmr/AlvianoDLR15} (version {\small\texttt{2.1}}), and tested the $6$ resulting configurations over the $6th$ ASP Competition benchmarks.

%

\begin{sidewaystable}
  \caption{\small $6th$ Competition (20 instances per problem) -- Solving Benchmarks: number of solved instances and average running times (in seconds). Time out and unsupported syntax issues are denoted by \texttt{TO} and \texttt{US}, respectively.}
  \tabcolsep=0.05cm
    \begin{tabular}{lrrrrrrrrrrrr}
    \hline\hline
    \multirow{2}[0]{*}{\textbf{Problem}} & \multicolumn{2}{c}{\textbf{\idlv$\vert$\clasp}} & \multicolumn{2}{c}{\textbf{\lpopt$\vert$\idlv$\vert$\clasp}} & \multicolumn{2}{c}{\textbf{\idlvsd$\vert$\clasp}} & \multicolumn{2}{c}{\textbf{\idlv$\vert$\wasp}} & \multicolumn{2}{c}{\textbf{\lpopt$\vert$\idlv$\vert$\wasp}} & \multicolumn{2}{c}{\textbf{\idlvsd$\vert$\wasp}} \\
    \cmidrule{2-3}     \cmidrule{4-5}      \cmidrule{6-7} \cmidrule{8-9}\cmidrule{10-11} \cmidrule{12-13}
    & \textbf{\#solved} & \textbf{time} & \textbf{\#solved} & \textbf{time} & \textbf{\#solved} & \textbf{time} & \textbf{\#solved} & \textbf{time} & \textbf{\#solved} & \textbf{time} & \textbf{\#solved} & \textbf{time} \\
    \hline
    Abstract Dialectical Frameworks & 20 & 6.88 & 20 & 7.36 & 20 & 6.89 & 11 & 33.26 & 11 & 22.38 & 11 & 32.21 \\
    Combined Configuration & 8  & 148.64 & 9  & 176.67 & 10 & 182.41 & 1  & 311.89 & 0  & TO & 0  & TO \\
    Complex Optimization & 18 & 149.84 & 19 & 167.58 & 18 & 149.44 & 6  & 150.30 & 5  & 99.05 & 6  & 148.00 \\
    Connected Still Life & 6  & 220.70 & 6  & 243.05 & 6  & 222.12 & 12 & 55.02 & 12 & 78.03 & 12 & 55.47 \\
    Consistent Query Answering & 20 & 87.05 & 0  & US & 20 & 87.39 & 18 & 87.47 & 0  & US & 18 & 88.05 \\
    Crossing Minimization & 7  & 53.02 & 6  & 64.23 & 7  & 56.79 & 19 & 3.50 & 19 & 2.40 & 19 & 5.58 \\
    Graceful Graphs & 9  & 141.77 & 10 & 130.97 & 9  & 140.44 & 6  & 174.17 & 4  & 122.14 & 6  & 171.42 \\
    Graph Coloring & 15 & 162.25 & 15 & 171.39 & 15 & 160.44 & 8  & 133.71 & 7  & 217.23 & 8  & 133.34 \\
    Incremental Scheduling & 13 & 90.62 & 11 & 39.16 & 14 & 128.81 & 8  & 155.20 & 5  & 137.74 & 6  & 124.89 \\
    Knight Tour With Holes & 11 & 55.76 & 10 & 26.90 & 11 & 56.04 & 10 & 35.50 & 8  & 63.97 & 10 & 35.67 \\
    Labyrinth & 12 & 63.19 & 11 & 120.09 & 12 & 67.25 & 11 & 104.73 & 10 & 168.65 & 11 & 106.71 \\
    Maximal Clique & 0  & TO & 0  & TO & 0  & TO & 9  & 353.03 & 9  & 353.63 & 9  & 352.56 \\
    MaxSAT & 7  & 39.67 & 7  & 46.81 & 7  & 39.77 & 19 & 91.01 & 19 & 95.82 & 19 & 91.49 \\
    Minimal Diagnosis & 20 & 8.90 & 20 & 8.46 & 20 & 8.32 & 20 & 30.77 & 20 & 29.38 & 20 & 25.95 \\
    Nomystery & 8  & 138.64 & 9  & 103.16 & 7  & 203.32 & 8  & 37.32 & 9  & 33.34 & 7  & 167.59 \\
    Partner Units & 14 & 19.81 & 14 & 20.29 & 14 & 20.26 & 5  & 116.99 & 10 & 168.07 & 10 & 168.24 \\
    Permutation Pattern Matching & 11 & 164.63 & 17 & 152.47 & 20 & 15.62 & 20 & 182.53 & 10 & 279.70 & 20 & 23.36 \\
    Qualitative Spatial Reasoning & 20 & 125.13 & 20 & 125.75 & 20 & 124.97 & 13 & 145.50 & 13 & 145.23 & 13 & 145.67 \\
    Reachability & 20 & 137.55 & 0  & US & 20 & 137.53 & 6  & 138.40 & 0  & US & 6  & 139.17 \\
    Ricochet Robots & 9  & 67.84 & 12 & 109.32 & 12 & 188.07 & 7  & 206.86 & 8  & 87.95 & 9  & 134.22 \\
    Sokoban & 8  & 73.95 & 9  & 82.45 & 8  & 76.90 & 8  & 86.00 & 9  & 64.59 & 8  & 88.25 \\
    Stable Marriage & 5  & 389.26 & 7  & 341.43 & 5  & 387.85 & 7  & 410.46 & 7  & 427.66 & 7  & 431.15 \\
    Steiner Tree & 3  & 243.89 & 3  & 244.89 & 3  & 242.45 & 1  & 131.66 & 1  & 131.75 & 1  & 131.80 \\
    Strategic Companies   & 17 & 119.63 & 0  & US & 17 & 122.24 & 7  & 31.38 & 0  & US & 7  & 30.95 \\
    System Synthesis & 0  & TO & 0  & TO & 0  & TO & 0  & TO & 0  & TO & 0  & TO \\
    Valves Location Problem & 16 & 43.09 & 16 & 26.09 & 16 & 43.05 & 15 & 40.93 & 15 & 39.27 & 15 & 41.32 \\
    Video Streaming & 13 & 61.84 & 10 & 75.70 & 13 & 61.63 & 9  & 9.15 & 0  & TO & 9  & 9.03 \\
    Visit-all & 8  & 16.90 & 8  & 15.22 & 8  & 15.21 & 8  & 62.11 & 8  & 61.28 & 8  & 60.06 \\
    \hline
    Total Solved Instances & \multicolumn{2}{c}{318/560} & \multicolumn{2}{c}{269/560} & \multicolumn{2}{c}{332/560} & \multicolumn{2}{c}{272/560} & \multicolumn{2}{c}{219/560} & \multicolumn{2}{c}{275/560} \\
    \hline\hline
    \end{tabular}%
  \label{table:solving}%
\end{sidewaystable}%

\begin{figure}
  \centering
  \includegraphics[width=\textwidth]{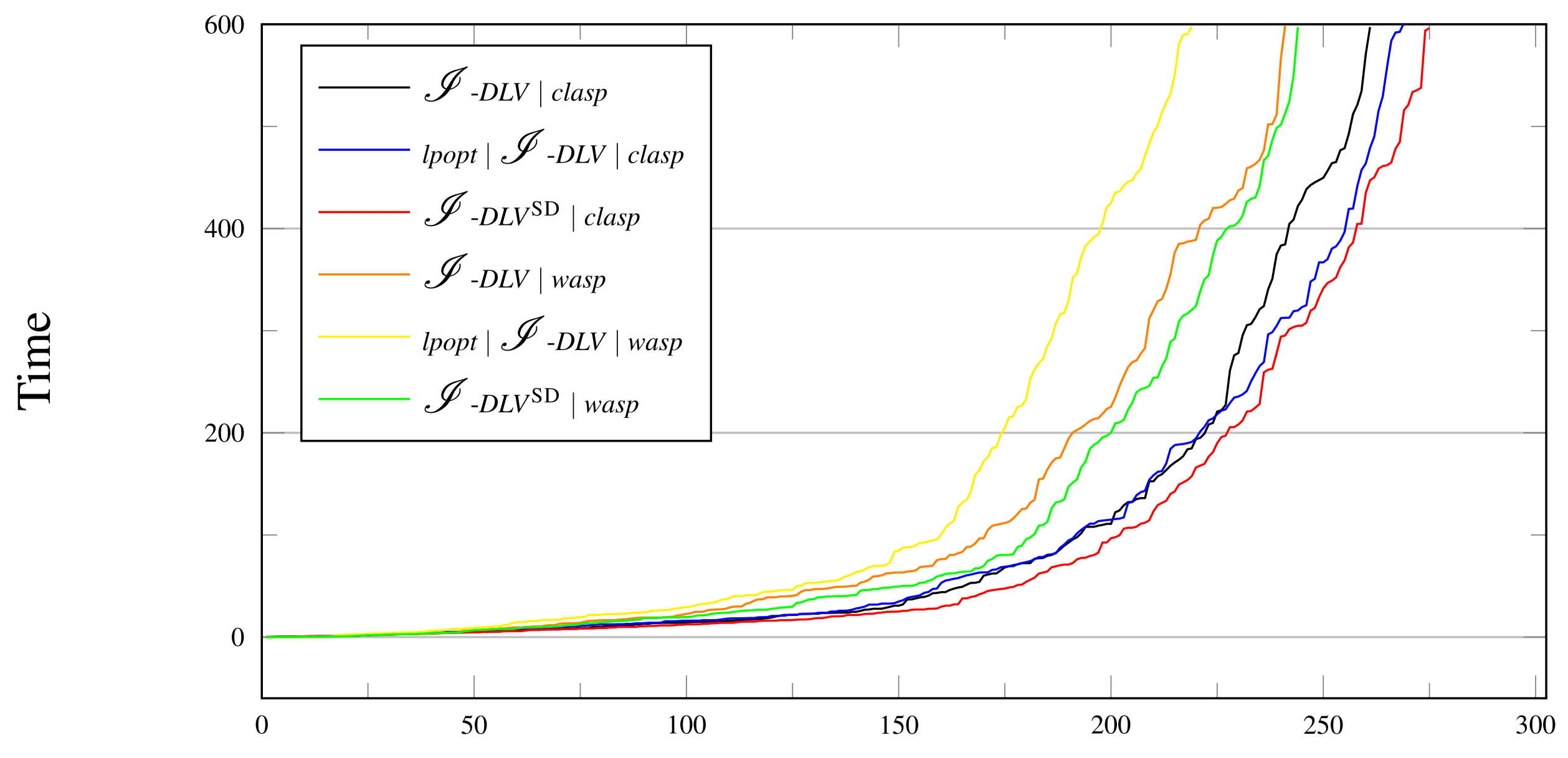}
  \caption{6th Competition Solving Benchmarks: solved instances over time (in seconds).\label{fig:6thCompSolvedbench1}}
\end{figure}

Average times and number of solved instances within the allotted time are reported in Table~\ref{table:solving}, where 
time outs and cases of unsupported syntax are denoted by \verb"TO" and \verb"US", respectively.
First of all, we observe that both solvers, when coupled with \idlvsd, show, in general, improved performance and solve a larger number of instances, w.r.t. the configurations with \idlv; on the contrary, the ``blind usage'' of \lpopt leads, in general, to a loss of performance for both solvers: in spite of the gain in some cases, the total number of solved instances within the suite is significantly lower.
A different perspective of the results is provided by Figure\re{fig:6thCompSolvedbench1}, showing solved instances over time (in seconds), where the benefits of the proposed techniques are very clear for both tested solvers.

Improvements on the overall ASP computation observable when \idlvsd is used are not only caused by improvements on grounding times; indeed, there are also cases in which solving times get better even if there is no evident gain in grounding times.
This is due to the fact that the rewriting causes changes in the ``form'' and the size of the generated instantiation, thus often inducing positive effects on the solving side.
Furthermore, it can be observed that on a same domain the effects of the decomposition on the two solvers are different: in some cases, benefits enjoyed by a solver are not reported for the other one (see, for instance, \emph{Incremental Scheduling}).
This suggests that the heuristics guiding the smart decomposition herein proposed, that already shows a general positive impact on both mainstream solvers, could be further fine-tuned once a specific solver to be coupled to \idlvsd is chosen, by taking into account also its specific characteristics.

%
%
\section{Conclusion}\label{sec:conclusion}
We introduced \decalg, a novel technique for automatically optimizing ASP programs by means of decomposition-guided rewritings.
The algorithm is designed to be adapted to different ASP implementations; furthermore, it can be customized  with heuristics of choice for discerning among possible
decompositions for each input rule, and determining whether applying the selected decomposition appears to be actually a ``smart'' choice.

In addition, we embedded a version of \decalg in the ASP system \dlv, and in particular in its grounding module \idlv.
We introduced heuristics criteria for selecting decompositions that consider not only the non-ground structure of the program at hand, but also the instance it is coupled to.
We experimentally tested our approach, and results are very promising: the proposed technique improves grounding performance, and highlights a positive impact, in general, also on the solving side.
This is confirmed also by the results of the 7th ASP
Competition\cit{DBLP:conf/lpnmr/GebserMR17}: here the winner was a system combining the version of \idlv implementing the preliminary decomposition rewriting described in\cit{calimeri-etal-paoasp-2017} with an automatic solver selector\cit{DBLP:conf/aiia/FuscaCZP17}, that inductively chooses the best solver depending on some inherent features of the instantiation produced.

\medskip

The \idlv system incorporating the technique herein described can be downloaded from \url{https://github.com/DeMaCS-UNICAL/I-DLV/wiki}, where a user guide is also reported addressing, among others, the options related to the techniques described in the present work.

\medskip

As future work, we plan to investigate on further strategies for generating decompositions, starting from a more fine-grained analysis along existing ones.
For instance, we note that current decompositions tend to split up a rule as much as possible, and in some cases this might require fresh predicates featuring significantly large extensions that could have a noticeable impact on performance;
hence, given a set of bags composing a tree decomposition, one could check whether collapsing some bags produces some benefits with this respect.
In addition, we also plan to take advantage from automatic and more advanced methods, such as machine learning mechanisms, in order to better tailor decomposition criteria and threshold values to the scenario at hand.
Furthermore, we want to design a version of \decalg specifically geared towards solvers, with the aim of further automatically optimizing the whole ASP computational process.
A starting point to this direction can be the recent work of~\cit{DBLP:conf/ijcai/BliemMMW17}, where it emerged that the performance of modern solvers are highly influenced by the tree-width of the input program; thus, this represents a starting point to explore the potential of our technique on the solving step.

\section*{Acknowledgements}\label{sec:ack}
This work has been partially supported by the Italian region Calabria under project
``DLV Large Scale'' (CUP J28C17000220006) POR Calabria FESR 2014--2020 and by both the European Union and the Italian Ministry of Economic Development under the project EU H2020 PON I\&C 2014--2020 ``Smarter Solutions in the Big Data World -- S2BDW'' (CUP B28I17000250008).

\bibliography{references-cleaned}


\appendix

\section{Experiments on Automatic Optimization}\label{appendix:sec:4thcomp}
In Section\re{sec:experiments} we discussed the results of tests over the benchmark suite from the $6th$ ASP Competition.
It is worth noting that many domains have been included in several subsequent editions of the ASP Competition series; over the years, the participant teams have iteratively fine-tuned the encodings with the aim of maximizing performance of competing ASP systems.
This led, in the case of $6th$ Competition, to a bunch of programs that are already optimized for ASP computation, thus limiting the room for further improvements.

\begin{table}
  \centering
  \caption{\small $4th$ Competition -- Grounding Benchmarks: number of grounded instances and average running times (in seconds). \texttt{US} indicates that corresponding configurations do not support the adopted syntax. \label{table:decomp2}}
    \tabcolsep=0.050cm
    \begin{tabular}{llrrrrrr}
    \hline\hline
    \multirow{2}[2]{*}{\textbf{Problem}} & \multicolumn{1}{c}{\multirow{2}[2]{*}{\textbf{\#inst.}}} & \multicolumn{2}{c}{\textit{\textbf{\idlv}}} & \multicolumn{2}{c}{\textit{\textbf{\lpopt $\vert$ \idlv}}} & \multicolumn{2}{c}{\textit{\textbf{\idlvsd}}} \\
    \cmidrule{3-4}     \cmidrule{5-6}      \cmidrule{7-8}
    & & \textbf{\#grounded} & \textbf{time} & \textbf{\#grounded} & \textbf{time} & \textbf{\#grounded} & \textbf{time} \\
    \hline
     Abstract Dialectical Frameworks & 30 & 30 & 0.13 & 30 & 0.13 & 30 & 0.13 \\
    Bottle Filling Problem & 30 & 30 & 4.12 & 30 & 6.86 & 30 & 4.39 \\
    Chemical Classification & 30 & 30 & 87.81 & 30 & 403.38 & 30 & 88.22 \\
    Complex Optimization * & 29 & 29 & 36.28 & 29 & 38.39 & 29 & 36.07 \\
    Connected Still Life * & 10 & 10 & 0.12 & 10 & 0.13 & 10 & 0.15 \\
    Crossing Minimization * & 30 & 30 & 0.10 & 30 & 0.10 & 30 & 0.10 \\
    Graceful Graphs & 30 & 30 & 0.37 & 30 & 0.39 & 30 & 0.37 \\
    Graph Colouring * & 30 & 30 & 0.10 & 30 & 0.10 & 30 & 0.10 \\
    Hanoi Tower & 30 & 30 & 0.22 & 30 & 0.23 & 30 & 0.30 \\
    Incremental Scheduling * & 30 & 12 & 297.95 & 17 & 229.49 & 21 & 221.17 \\
    Knight Tour with Holes * & 30 & 20 & 176.99 & 20 & 181.16 & 20 & 178.59 \\
    Labyrinth & 30 & 30 & 1.49 & 30 & 1.40 & 30 & 1.51 \\
    Maximal Clique * & 30 & 30 & 0.34 & 30 & 1.11 & 30 & 0.34 \\
    Minimal Diagnosis * & 30 & 30 & 2.54 & 30 & 2.20 & 30 & 2.57 \\
    Nomystery * & 30 & 30 & 34.91 & 21 & 100.14 & 30 & 35.28 \\
    Permut. Pattern Matching * & 30 & 28 & 57.71 & 30 & 3.64 & 30 & 62.32 \\
    Qualitative Spatial Reasoning * & 30 & 30 & 2.85 & 30 & 2.87 & 30 & 2.84 \\
    Reachability & 30 & 30 & 101.93 & 0  & US & 30 & 102.04 \\
    Ricochet Robots & 30 & 30 & 0.27 & 30 & 0.31 & 30 & 0.31 \\
    Sokoban & 30 & 30 & 2.65 & 30 & 2.68 & 30 & 2.69 \\
    Solitaire & 27 & 27 & 0.13 & 27 & 0.18 & 27 & 0.20 \\
    Stable Marriage * & 30 & 30 & 28.35 & 30 & 2.65 & 30 & 2.46 \\
    Strategic Companies & 30 & 30 & 0.19 & 0  & US & 30 & 0.19 \\
    Valves Location & 30 & 30 & 3.97 & 30 & 3.98 & 30 & 3.93 \\
    Visit-all * & 30 & 30 & 0.13 & 30 & 0.14 & 30 & 0.13 \\
    Weighted-Sequence Problem * & 30 & 30 & 2.87 & 30 & 9.61 & 30 & 2.95 \\
    \hline
    Total Grounded Instances & & \multicolumn{2}{c}{726/756} & \multicolumn{2}{c}{664/756} & \multicolumn{2}{c}{737/756} \\
    \hline\hline
    \end{tabular}
\end{table}

On the one hand, such considerations strengthen the positive results reported in Section\re{sec:experiments}; on the other hand, one might wonder about what is the impact of the \decalg algorithm when dealing with less refined encodings.
We find such inquiry of interest, as it should be in the spirit of the declarative nature of ASP to allow developers to concentrate on knowledge representation rather than on low-level performance issues, that also lead, in some cases, to the production of encodings rather involved and less ``human readable''.
This is why, in addition to the experiments reported in Section\re{sec:experiments}, we tested our approach also on benchmarks coming from older editions of the ASP Competition series.
In particular, we considered the $4th$ Competition, as it is the farthest in time in which encodings comply to the \aspcore input language standard.

In this set of experiments we measured grounding times of the same three versions of \idlv already taken into account in Section\re{sec:experiments}, within the same experimental environment (hardware, software, memory and time limits).
Table\re{table:decomp2} shows the results; problem names reported in italic denotes domains which are in common between the $4th$ and the $6th$ Competition, while those marked with a '$*$' symbol feature more optimized encodings in the $6th$. 
%
%
As expected, in this scenario the positive impact of \decalg is even more evident: \idlvsd grounds a larger number of instances in a significant smaller average time.
Intuitively, the less an encoding is fine-tuned, the highest are likely to be the benefits stemming from a careful automatic rewriting of input rules.

\begin{table}
  \caption{\small Variation of the encodings -- Grounding Benchmarks: number of grounded instances and average running times (in seconds).  \label{table:common}}
  \centering
    \tabcolsep=0.020cm
    \begin{tabular}{llrrrrrrrr}
    \hline\hline
    \textbf{Problem} & \multicolumn{1}{l}{\textbf{\#inst.}} & \multicolumn{4}{c}{\textit{\textbf{\idlv}}} & \multicolumn{4}{c}{\textit{\textbf{\idlvsd}}} \\
    \hline
    & & \multicolumn{1}{c}{\textbf{\#grounded}} & \multicolumn{1}{c}{\textbf{time}} & \multicolumn{1}{c}{\textbf{\#grounded}} & \multicolumn{1}{c}{\textbf{time}} & \multicolumn{1}{c}{\textbf{\#grounded}} & \multicolumn{1}{c}{\textbf{time}} & \multicolumn{1}{c}{\textbf{\#grounded}} & \multicolumn{1}{c}{\textbf{time}} \\
    \hline
    \multicolumn{10}{c}{$4th$ Competition Instances} \\
    \hline
    & & \multicolumn{2}{c}{\textbf{$4th$ Comp. Enc.}} & \multicolumn{2}{c}{\textbf{$6th$ Comp. Enc.}} & \multicolumn{2}{c}{\textbf{$4th$ Comp. Enc.}} & \multicolumn{2}{c}{\textbf{$6th$ Comp. Enc.}} \\
    \hline
    Incr. Scheduling & 30 & 12 & 297.95 & 30 & 54.65 & 21 & 221.17 & 30 & 1.93 \\
    Maximal Clique & 30 & 30 & 0.34 & 30 & 2.96 & 30 & 0.34 & 30 & 3.11 \\
    Minimal Diagnosis & 30 & 30 & 2.54 & 30 & 1.76 & 30 & 2.57 & 30 & 1.76 \\
    Nomystery & 30 & 30 & 34.91 & 30 & 47.24 & 30 & 35.28 & 30 & 47.11 \\
    Perm. Pattern Match. & 30 & 28 & 57.71 & 30 & 0.27 & 30 & 62.32 & 30 & 0.27 \\
    Stable Marriage & 30 & 30 & 28.35 & 30 & 3.16 & 30 & 2.46 & 30 & 2.86 \\
    \hline
    \multicolumn{10}{c}{$6th$ Competition Instances} \\
    \hline
    & & \multicolumn{2}{c}{\textbf{$4th$ Comp. Enc.}} & \multicolumn{2}{c}{\textbf{$6th$ Comp. Enc.}} & \multicolumn{2}{c}{\textbf{$4th$ Comp. Enc.}} & \multicolumn{2}{c}{\textbf{$6th$ Comp. Enc.}} \\
    \hline
    Incr. Scheduling & 20 & 11 & 336.77 & 20 & 16.07 & 19 & 211.61 & 20 & 16.21 \\
    Maximal Clique & 20 & 20 & 6.63 & 20 & 4.93 & 20 & 6.58 & 20 & 4.96 \\
    Minimal Diagnosis & 20 & 20 & 4.12 & 20 & 5.09 & 20 & 4.14 & 20 & 4.22 \\
    Nomystery & 20 & 20 & 55.11 & 20 & 3.45 & 20 & 43.74 & 20 & 3.63 \\
    Perm. Pattern Match. & 20 & 16 & 168.93 & 20 & 130.47 & 20 & 150.99 & 20 & 4.21 \\
    Stable Marriage & 20 & 0  & TO & 20 & 118.55 & 20 & 172.68 & 20 & 119.53 \\
    \hline\hline
    \end{tabular}%
\end{table}

Furthermore, for all problems in common between the two ASP competitions herein considered, we tested the systems over the programs obtained by coupling the encodings featured by the $4th$ with the instances featured by the $6th$, and vice-versa.
%
This should provide us with further information about a the impact of both ``manual'' optimizations and the ones coming from our automatic method.
Intuitively, an encoding may be optimized in different ways, and not necessarily by means of a syntactic modification of rules; for instance, one can push additional information about the domain at hand into the encoding, possibly with constraints, in order to reduce the search space.  
The results are reported in Table\re{table:common}. 
It is clear that, as one can expect, the best combination is given by \idlvsd fed with optimized encoding coming from $6th$ Competition.
However, some interesting considerations can be made: indeed, in several cases (for instance, {\em Permutation Pattern Marching}, both over instances from the $4th$, and, even more, over instances from the $6th$, that appear to be harder), the smart decomposition guarantees similar or better performance improvements in grounding times that are obtained by the manual tuning.
\section{Additional Experiments}\label{appendix:sec:alphaappendix}
\begin{table}
  \centering
  \caption{\small Additional Grounding Benchmarks: number of grounded instances and average running times (in seconds).}
    \begin{tabular}{llcccccc}
    \hline\hline
    \multirow{2}[2]{*}{\textbf{Problem}} & \multicolumn{1}{c}{\multirow{2}[2]{*}{\textbf{\#inst.}}} & \multicolumn{2}{c}{\textit{\textbf{\idlv}}} & \multicolumn{2}{c}{\textit{\textbf{\lpopt $\vert$ \idlv}}} & \multicolumn{2}{c}{\textit{\textbf{\idlvsd}}} \\
    \cmidrule{3-4}     \cmidrule{5-6}      \cmidrule{7-8}
    & & \textbf{\#grounded} & \textbf{time} & \textbf{\#grounded} & \textbf{time} & \textbf{\#grounded} & \textbf{time} \\
    \hline
    Cutedge & 130 & \multicolumn{1}{r}{130} & \multicolumn{1}{r}{34.38} & \multicolumn{1}{r}{130} & \multicolumn{1}{r}{1.64} & \multicolumn{1}{r}{130} & \multicolumn{1}{r}{1.08} \\
    Graph 5col & 180 & \multicolumn{1}{r}{180} & \multicolumn{1}{r}{0.12} & \multicolumn{1}{r}{180} & \multicolumn{1}{r}{0.14} & \multicolumn{1}{r}{180} & \multicolumn{1}{r}{0.12} \\
    Ground Explosion 2 & 17 & \multicolumn{1}{r}{7} & \multicolumn{1}{r}{73.33} & \multicolumn{1}{r}{7} & \multicolumn{1}{r}{73.02} & \multicolumn{1}{r}{7} & \multicolumn{1}{r}{72.87} \\
    Reach & 50 & \multicolumn{1}{r}{50} & \multicolumn{1}{r}{0.29} & \multicolumn{1}{r}{50} & \multicolumn{1}{r}{0.76} & \multicolumn{1}{r}{50} & \multicolumn{1}{r}{0.30} \\
    TimeTabling & 27 & \multicolumn{1}{r}{27} & \multicolumn{1}{r}{85.57} & \multicolumn{1}{r}{27} & \multicolumn{1}{r}{63.67} & \multicolumn{1}{r}{27} & \multicolumn{1}{r}{57.67} \\
    \hline
    \multicolumn{2}{l}{Total Solved Instances} & \multicolumn{2}{c}{\textbf{367/377}} & \multicolumn{2}{c}{\textbf{367/377}} & \multicolumn{2}{c}{\textbf{367/377}} \\
    \hline\hline
    \end{tabular}%
  \label{tab:alphabench}%
\end{table}%

We report next the results of an additional experimental evaluation over a further set of benchmark.
We take into account problem domains that have been already used for assessing performance of ASP systems in other works; in particular, we considered the domains \textit{Cutedge}, \textit{Graph 5col}, \textit{Ground Explosion 2}, \textit{Reach}~\cite{DBLP:conf/lpnmr/Weinzierl17}
and \textit{TimeTabling}~\cite{DBLP:journals/amai/PerriSCL07,DBLP:journals/jal/CalimeriPR08,DBLP:journals/tplp/PerriRS13}.

Table~\ref{tab:alphabench} reports grounding times of the same three versions of \idlv already taken into account in Section~\ref{sec:experiments}, within the same experimental environment (hardware, software, memory and time limits).
We first note that, regarding \textit{Reach}, even if the problem domain is the same as \textit{Reachability} of Section~\ref{sec:experiments}, both encoding and instances are different, as they are taken from~\cite{DBLP:conf/lpnmr/Weinzierl17}, and do not feature queries.
In this case, where decomposition is not applicable because of the rule structure, results show that the use of \decalg, as previously noted, allows us to avoid the overhead due to the invocation of \lpopt.
On the overall, again, one can observe the positive impact of \decalg over grounding times, that allows significant performance improvements in case of \textit{Cutedge} and \textit{TimeTabling}, where \idlvsd times are $97$\% and $33$\% lower, respectively, w.r.t. \idlv.

\end{document}